\documentclass{article}

\usepackage{abstract}
\usepackage{arxiv}
\usepackage[utf8]{inputenc}
\usepackage[T1]{fontenc}
\usepackage{hyperref}
\usepackage{url}
\usepackage{booktabs}
\usepackage{amsfonts}
\usepackage{nicefrac}
\usepackage{lipsum}
\usepackage{graphicx}
\usepackage{algorithmicx}
\usepackage[numbers]{natbib}
\usepackage{doi}
\usepackage{subcaption}
\usepackage{amssymb}
\usepackage{float}
\usepackage{amsmath}
\usepackage{setspace}
\usepackage{multirow}
\usepackage{svg}
\usepackage{fancyhdr}
\usepackage{hyperref}
\usepackage{algorithm}
\usepackage{algorithmicx}
\usepackage{algpseudocode}
\title{OralBBNet: Spatially Guided Dental Segmentation of Panoramic X-Rays with Bounding Box Priors}

\author{
\centering
\begin{tabular}{c c}
\href{https://orcid.org/0009-0009-6353-9808}{\includegraphics[scale=0.06]{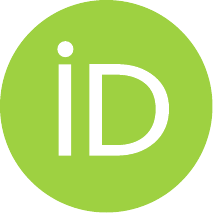}\hspace{1mm}Devichand Budagam} & 
\href{https://orcid.org/0000-0003-3719-4091}{\includegraphics[scale=0.06]{orcid.pdf}\hspace{1mm}Azamat Zhanatuly Imanbayev} \\
Department of Computer Science and Engineering & School of IT and Engineering \\
Indian Institute of Technology Kharagpur & Kazakh-British Technical University \\
Kharagpur, India & Almaty, Kazakhstan \\
\texttt{devichand579@kgpian.iitkgp.ac.in} & \texttt{a.imanbaev@kbtu.kz} \\[1em]
\href{https://orcid.org/0000-0002-3221-9352}{\includegraphics[scale=0.06]{orcid.pdf}\hspace{1mm}Iskander Rafailovich Akhmetov} & 
\href{https://orcid.org/0000-0001-9869-4909}{\includegraphics[scale=0.06]{orcid.pdf}\hspace{1mm}Aleksandr Sinitca} \\
School of IT and Engineering & Intelligent Devices Institute \\
Kazakh-British Technical University & St. Petersburg Electrotechnical University "LETI" \\
Almaty, Kazakhstan & St. Petersburg, Russia \\
\texttt{i.akhmetov@kbtuedu.onmicrosoft.com} & \texttt{amsinitca@etu.ru} \\[1em]
\href{https://orcid.org/0000-0003-2765-4509}{\includegraphics[scale=0.06]{orcid.pdf}\hspace{1mm}Sergey Antonov} & 
\href{https://orcid.org/0000-0003-2765-4509}{\includegraphics[scale=0.06]{orcid.pdf}\hspace{1mm}Dmitrii Kaplun*}\\
Department of Automation and Control Processes & Intelligent Devices Institute \\
St. Petersburg Electrotechnical University "LETI" & St. Petersburg Electrotechnical University "LETI" \\
St. Petersburg, Russia & St. Petersburg, Russia \\
\texttt{saantonov@etu.ru} & \texttt{*Corresponding author: dikaplun@etu.ru} \\[1em]
 \\
\end{tabular}
}

\date{}

\renewcommand{\headeright}{}
\renewcommand{\shorttitle}{}

\hypersetup{
pdftitle={Teeth Segmentation},
pdfsubject={q-bio.NC, q-bio.QM},
pdfauthor={Aleksandr M.~Sinitca, Dmitrii I.~Kaplun},
pdfkeywords={First keyword, Second keyword, More},
}

\begin{document}
\maketitle
\begin{abstract}
Teeth segmentation and recognition play a vital role in a variety of dental applications and diagnostic procedures. The integration of deep learning models has facilitated the development of precise and automated segmentation methods. Although prior research has explored teeth segmentation, not many methods have successfully performed tooth segmentation and detection simultaneously. This study presents UFBA-425, a dental dataset derived from the UFBA-UESC dataset, featuring bounding box and polygon annotations for \(425\) panoramic dental X-rays. In addition, this paper presents the OralBBNet architecture, which is based on the best segmentation and detection qualities of architectures such as U-Net and YOLOv8, respectively. OralBBNet is designed to improve the accuracy and robustness of tooth classification and segmentation on panoramic X-rays by leveraging the complementary strengths of U-Net and YOLOv8. Our approach achieved a \(1-3\%\) improvement in mean average precision (mAP) for tooth detection compared to existing techniques and a \(15-20\%\) improvement in the dice score for teeth segmentation over state-of-the-art (SOTA) solutions for various tooth categories and \(2-4\%\) improvement in the dice score compared to other SOTA segmentation architectures. The results of this study establish a foundation for the wider implementation of object detection models in dental diagnostics.
\end{abstract}
\section{Introduction}
\label{sec: introduction}
\subsection{Motivation}
The demand for dental care and qualified dentists is growing due to several factors, including population growth, increasing life expectancy, and a greater emphasis on oral health. As the field of dentistry evolves, advanced technologies are required to improve diagnostic accuracy, optimize treatment planning, and improve patient care. Deep learning architectures have emerged as a promising solution for dental image analysis, offering significant potential for automation and efficiency gains \cite{arsiwala2023machine, doi:10.1142/S2737599423300015}. Despite their achievements in numerous medical imaging applications, deep learning models continue to encounter challenges due to misinterpretations of results and their inability to offer detailed information.\\[5pt]
Teeth segmentation and classification are essential in dental imaging, consisting of two main tasks: first, precisely outlining each tooth by assigning image pixels to corresponding anatomical features, and second, numbering the teeth according to standardized dental systems. These tasks play a vital role in numerous applications such as dental diagnostics, orthodontic treatment planning, and forensic identification. Traditionally, manual segmentation and annotation of teeth are performed by dental specialists. This method is not only time-consuming and labor-intensive but also prone to inconsistencies due to variations in image quality, anatomical differences among patients, and observer subjectivity \cite{Zheng2024}. Furthermore, panoramic X-rays, which are widely used in dental diagnostics, present additional challenges such as high variability in tooth shape and positioning, low contrast, and noise artifacts \cite{doi:10.1177/09544119231217603}. These factors make automated segmentation a difficult task, as standard computer vision techniques may struggle to accurately detect and classify teeth in such complex images. Misclassifications or segmentation errors can lead to incorrect diagnoses and treatment plans. A key challenge in developing deep learning techniques for dental imaging is the scarcity of high-quality annotated datasets. Unlike radiology or dermatology, dental datasets are limited, fragmented, and often restricted by privacy laws. This scarcity hampers the training of deep learning models, which need extensive data to generalize across various patients and conditions \cite{10.1093/dmfr/twad001}. Manual annotation is time-consuming and requires dental expertise, limiting large-scale dataset creation and hindering new technique development. Insufficient data leads to overfitting and poor model performance, making it crucial to address this scarcity for progress in automated teeth segmentation and classification.\\[5pt]
Given these challenges, automated deep learning-based teeth segmentation and classification have gained increasing attention due to their success in handling complex computer vision tasks. However, existing approaches still face difficulties in precisely recognizing and localizing each tooth due to the aforementioned issues. Thus, improving the robustness and accuracy of automated segmentation models remains a critical research challenge.
\subsection{Contributions}
This work presents OralBBNet, an enhanced model for classifying teeth and performing instance segmentation, which incorporates spatial prior knowledge into a U-Net \cite{ronneberger2015u} framework. By employing a one-stage detection module, YOLO \cite{redmon2016you}, to capture spatial features, we improve both efficiency and accuracy of the segmentation, moving away from traditional two-stage object detection models like Mask R-CNN. The major contributions of our work are:
\begin{enumerate}
    \item Created UFBA-425, a new public dental dataset derived from UFBA-UESC \cite{Jader2018AutomaticST}. It is one of the largest publicly available datasets, with annotations for segmentation and classification. Featured in the Roboflow 100-VL benchmark \cite{robicheaux2025roboflow100} and considered challenging, requiring strong contextual and spatial understanding for teeth classification and segmentation.
    \item OralBBNet, a new segmentation architecture, was developed to perform both teeth numbering and segmentation  with improved spatial prior knowledge.
    \item Comprehensive experiments and comparative studies are conducted to evaluate the model's robustness and the dataset's complexity.
\end{enumerate}
The rest of the paper is organized as follows: Section \ref{sec: related work} covers the literature related to tooth segmentation and detection. Section \ref{sec: methodology} describes the dataset creation, model architecture, and training pipeline. Section \ref{sec: results} provides insights into the experimental setup and the results achieved. Section \ref{sec: ablations} analyzes the comparative study alongside other current teeth segmentation and detection frameworks. Section \ref{sec: conclusion} concludes the paper. Section \ref{sec: limitations} outlines the limitations of the study.
\section{Related Work}
\label{sec: related work}
\textbf{Teeth Segmentation and Numbering:} In their study, Pinheiro et al. \cite{2021SPIE12088E..0CP} developed a technique for numbering both permanent and deciduous teeth using deep instance segmentation on panoramic X-rays. tackling issues like overlapping tooth instances and diverse tooth structures. They utilized Mask R-CNN with various segmentation heads, including PointRend \cite{19} and FCN \cite{20}, and achieved good results. Meanwhile, Indraswari et al. \cite{Indraswari201649} suggested a method for segmenting teeth in low-contrast panoramic radiographs through a three-step process: initially generating vertical and horizontal directional images via \textit{Decimation-Free Directional Filter Bank Thresholding}, then enhancing these images to highlight tooth edges and minimize noise, followed by applying \textit{Multistage Adaptive Thresholding} combined with \textit{Sauvola Local Thresholding} for segmentation. Their experiments on \(40\) tooth images showed this method surpassed other thresholding methods. Likewise, Silva et al. \cite{Silva2020ASO} conducted research on both tooth segmentation and numbering using end-to-end deep neural networks and investigated various deep learning architectures, such as PANet, HTC, ResNeSt, and Mask R-CNN, thus demonstrating the capability of deep learning models in automating these tasks. TSegNet \cite{cui2021tsegnet}, an efficient and accurate tooth segmentation network for 3D dental models, employs a two-stage network framework. Recently, Beser et al. \cite{Beser2024} presented a deep learning method utilizing YOLOv5 to automatically detect, segment, and number teeth in pediatric patients with mixed dentition using panoramic radiographs.

Koch et al. \cite{Koch2019AccurateSO} utilized U-Nets for precise segmentation of dental panoramic radiographs, showing the model's competency in managing complex dental structures and variations in radiographic imagery. Building on this, \textit{Two-Stage Attention Segmentation Network} (TSASNet) \cite{Zhao2020TSASNetTS} was introduced for tooth segmentation in dental panoramic X-rays and enhances segmentation accuracy by concentrating on important regions. Furthermore, \textit{Multi-Scale Location Perception Network} (MSLPNet) \cite{Chen2021MSLPNetML} developed for segmenting dental panoramic X-rays, which employs multi-scale feature extraction to better capture detailed dental structures. This method addresses the difficulties of varying tooth sizes and orientations in panoramic images. Jader et al. \cite{Jader2018DeepIS} leveraged deep instance segmentation methods to precisely identify and segment individual teeth in panoramic X-ray images, supporting more accurate dental evaluations. Lee et al. \cite{app12010475} developed a deep neural network to automatically detect mandibular third molars in panoramic radiographic images and predict both extraction difficulty and the likelihood of inferior alveolar nerve (IAN) injury. Tekin et al. \cite{YARENTEKIN2022105547} improved the segmentation and numbering of teeth based on FDI notation in bitewing radiographs utilizing convolutional neural networks, achieving notable precision and mAP scores. Zhao et al. \cite{ZHAO2023102447} used the Mask R-CNN algorithm to recognize and segment teeth and mandibular nerve canals in panoramic dental X-rays, successfully identifying each tooth, including any missing ones, as well as the mandibular nerve canals, thus addressing the challenges posed by complex oral structures in these radiographs. Meanwhile, Teeth U-Net \cite{HOU2023106296}, a segmentation model tailored for dental panoramic X-rays, integrates context semantics and contrast enhancement to boost segmentation accuracy and support clinical diagnoses.

\textbf{Maxillofacial Region Segmentation:}  Kong et al. \cite{kong2020automated} introduced an efficient encoder-decoder network for automated maxillofacial segmentation in panoramic dental X-ray images, demonstrating high accuracy in segmenting maxillofacial structures and improving diagnostic precision. Additionally, traditional approaches such as the active contour model have been explored for segmentation tasks. Divya et al. \cite{divya2016appending} applied an active contour model to digital panoramic dental X-ray images, improving segmentation performance by refining region boundaries.

\textbf{Dental Caries Detection:} PaxNet \cite{haghanifar2020paxnet}, a model leveraging ensemble transfer learning and capsule networks for detecting dental caries in panoramic X-rays, enhancing detection accuracy through pre-trained models and capsule classifiers. Similarly, Singh et al. \cite{singh2021gv} proposed an optimal CNN-LSTM classifier for GV Black dental caries classification and preparation techniques, improving diagnostic precision. Wang et al. \cite{wang2020automated} developed an automated classification framework using dual-channel dental imaging with convolutional neural networks (CNNs) to analyze auto-fluorescence and white light images, enabling more accurate caries detection. Furthermore, Xu et al. \cite{XU2022102119} introduced an AI-assisted method for identifying the history of root canal therapy from periapical films, utilizing SIFT-SVM, CNN, and transfer learning.

Table \ref{tab:1} summarizes the related frameworks, highlighting how the application of deep learning in dental image analysis has significantly improved performance across various diagnostic tasks. Continuous research and technological advancements are further enhancing the accuracy and efficiency of these automated systems, ultimately supporting improved patient outcomes in dental care.
\section{Materials and Method}
\label{sec: methodology}
\subsection{UFBA-425 Dataset Construction}
In dental image analysis, achieving high prediction accuracy is critically dependent on the availability of comprehensive, annotated datasets. However, there is a notable scarcity of such datasets, particularly for tasks like tooth
\begin{table}
    \centering
    \renewcommand{\arraystretch}{1.4} 
    \setlength{\tabcolsep}{1.5pt} 
    \resizebox{\linewidth}{!}{ 
    \begin{tabular}{l l l l l}
    \toprule
       Reference & Radiograph Type & Functionality & Objective & Algorithm \\
    \midrule
        \cite{app12010475}  & Panoramic image  & Mandibular nerve detection & Detection & Vision Transformer \\
        \cite{XU2022102119} & Periapical film & Detect caries & Detection & Faster R-CNN \\
        \cite{haghanifar2020paxnet} & Panoramic image & Detect caries & Detection & PaxNet \\
        \cite{singh2021gv}  & Periapical film & Classification of dental caries grade & Classification & CNN + LSTM \\
        \cite{wang2020automated} & Dual channel image & Classification of early-stage caries & Classification & CNN \\
        \cite{divya2016appending} & Panoramic image & Segmentation of the maxillofacial region & Segmentation & Active Contour Model \\
        \cite{kong2020automated} & Panoramic image & Segmentation of the maxillofacial region & Segmentation & EED-Net \\
        \cite{Indraswari201649} & Panoramic image & Teeth segmentation & Segmentation  & Filter Bank Thresholding \\
        \cite{Koch2019AccurateSO} & Panoramic image & Teeth segmentation & Segmentation & U-Net \\
        \cite{Zhao2020TSASNetTS} & Panoramic image & Teeth segmentation & Segmentation & TSASNet \\
        \cite{Chen2021MSLPNetML} & Panoramic image & Teeth segmentation & Segmentation & MSLPNet \\
        \cite{Jader2018DeepIS} & Panoramic image & Teeth segmentation & Segmentation & Mask R-CNN \\
        \cite{2021SPIE12088E..0CP} & Panoramic image & Instance Segmentation of teeth & Segmentation & Mask R-CNN \\
        \cite{Silva2020ASO} & Panoramic image & Instance Segmentation and numbering of teeth & Segmentation & PANet, HTC, ResNeSt, Mask R-CNN \\
        \cite{cui2021tsegnet} & CBCT image & Instance Segmentation of teeth & Segmentation & TsegNet \\
        \cite{YARENTEKIN2022105547} & Bitewing radiograph & Instance Segmentation of teeth & Segmentation & Mask R-CNN \\
        \cite{ZHAO2023102447} & Panoramic image & Segmentation of teeth and mandibular nerve canals & Segmentation & Mask R-CNN \\
        \cite{HOU2023106296} & Panoramic image & Instance Segmentation of teeth & Segmentation & Teeth U-Net \\
        \cite{Beser2024} & Panoramic image & Instance Segmentation and numbering of teeth & Segmentation and Detection & YOLO v5 \\
        This work & Panoramic image & Instance Segmentation and numbering of teeth & Segmentation and Detection & OralBBNet \\
    \bottomrule
    \end{tabular}
    } 
    \vspace{5pt} 
    \caption{Deep learning research on dental images for different functionalities and objectives.}
    \label{tab:1}
\end{table}
numbering and instance segmentation. Our preparatory work addresses this gap by carefully selecting and processing a representative dataset for subsequent model training. While initiatives like the DENTEX challenge \cite{hamamci2023dentexabnormaltoothdetection} have made strides by providing datasets with detection labels, there remains a dearth of publicly available datasets specifically tailored for instance segmentation tasks.
This scarcity poses challenges for researchers aiming to develop and validate advanced segmentation algorithms in dental image analysis. Our work contributes to addressing this gap by offering a meticulously annotated subset of the UFBA-UESC dataset \cite{Jader2018AutomaticST}, thereby facilitating further research and development in this critical area.\\[5pt]
We based our study on the UFBA-UESC dataset, an extensive collection of anonymized panoramic X-ray dental images exhibiting high variability. This dataset comprises \(1500\) images categorized into \(10\) distinct classes, reflecting various dental conditions such as standard \(32\) teeth, the presence of dental appliances, and dental restorations. It also includes images with fewer than \(32\) teeth due to extractions and cases with supernumerary teeth resulting from abnormal mutations. This diversity mirrors real-world variations in dental scans, encompassing factors like dental anomalies and missing teeth. The structure of the UFBA-UESC dataset is detailed in Table \ref{tab:2}. A significant limitation of the UFBA-UESC dataset is that the image annotations required for training models in tooth detection and numbering tasks are not publicly available. To mitigate this, we manually annotated \(425\) X-ray images from the dataset, ensuring a representative distribution across all categories. These annotations encompass both instance segmentation and object detection labels, providing a foundational dataset for training and evaluating our models.
\subsection{Annotation Policy}
\label{sec: annotation}
\begin{figure}
    \centering
    \begin{subfigure}{0.24\linewidth}
        \centering
        \includegraphics[width=\linewidth]{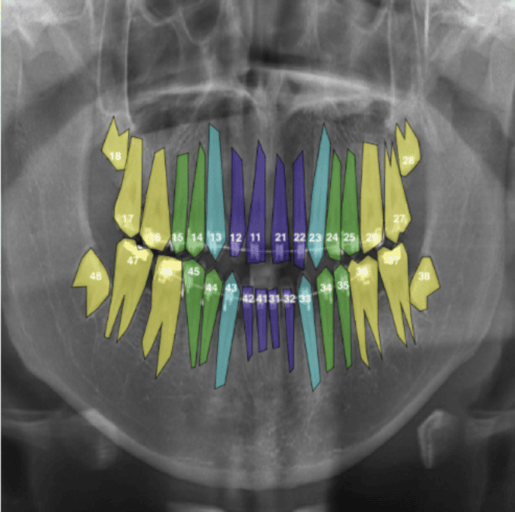}
        \label{fig:1a}
    \end{subfigure}
    \begin{subfigure}{0.24\linewidth}
        \centering
        \includegraphics[width=\linewidth]{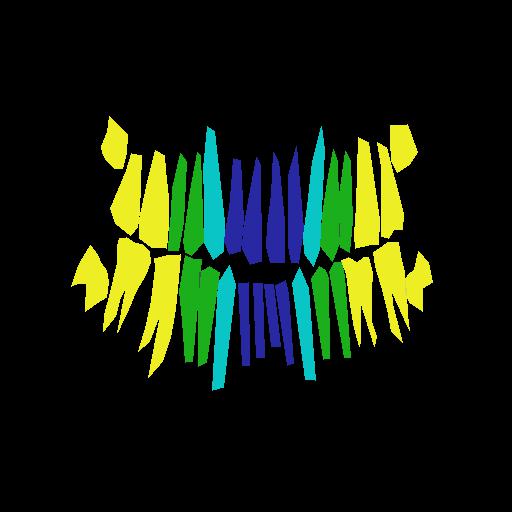}
        \label{fig:1b}
    \end{subfigure}
    \begin{subfigure}{0.24\linewidth}
        \centering
        \includegraphics[width=\linewidth]{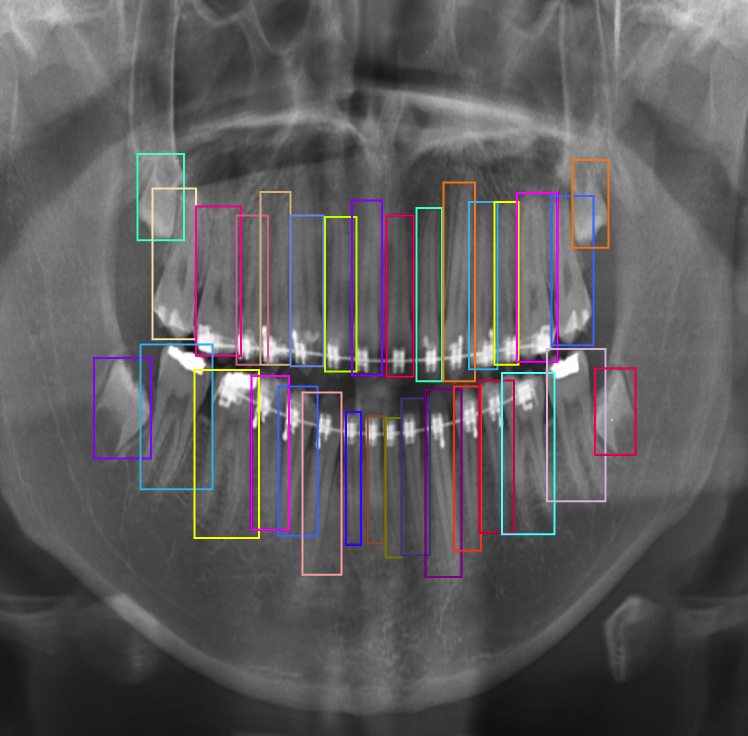}
        \label{fig:1c}
    \end{subfigure}
    \begin{subfigure}{0.24\linewidth}
        \centering
        \includegraphics[width=\linewidth]{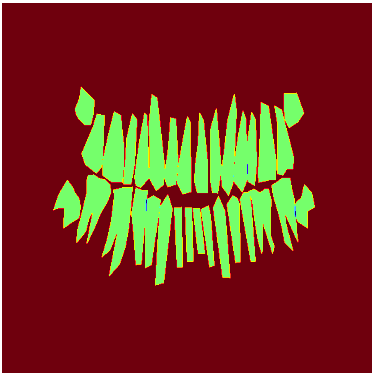}
        \label{fig:1d}
    \end{subfigure}
    \caption{(a) FDI notation system (b) Annotated polygon mask (c) Annotated bounding Box scan (d) Binary mask of polygon-based annotation}
    \label{fig:1}
\end{figure}

\begin{figure}
    \centering
    \begin{subfigure}{0.45\linewidth}
        \centering
        \includegraphics[width=\linewidth]{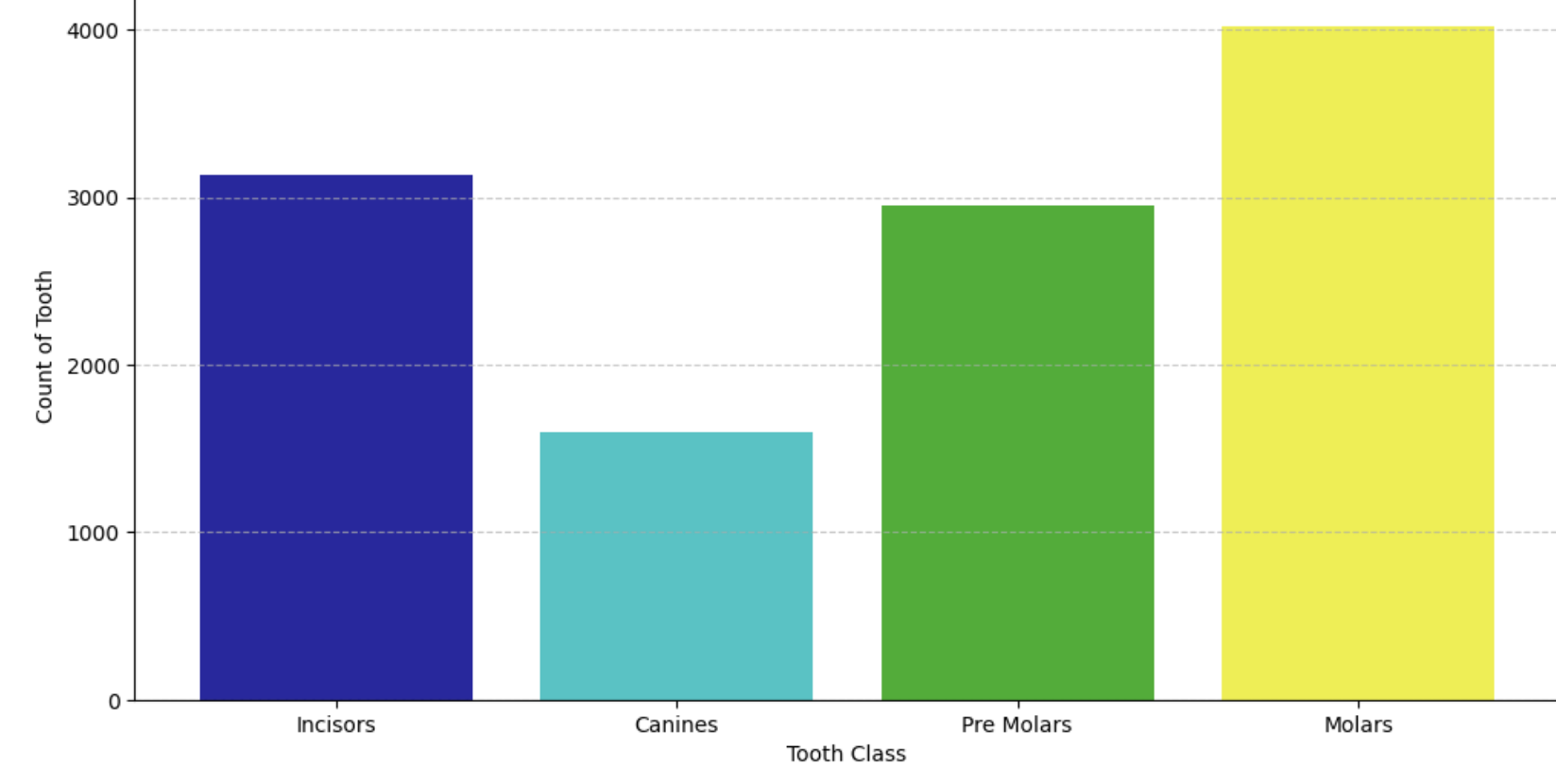}
        \caption{}
        \label{fig:subfiga}
    \end{subfigure}
    \hfill
    \begin{subfigure}{0.45\linewidth}
        \centering
        \includegraphics[width=\linewidth]{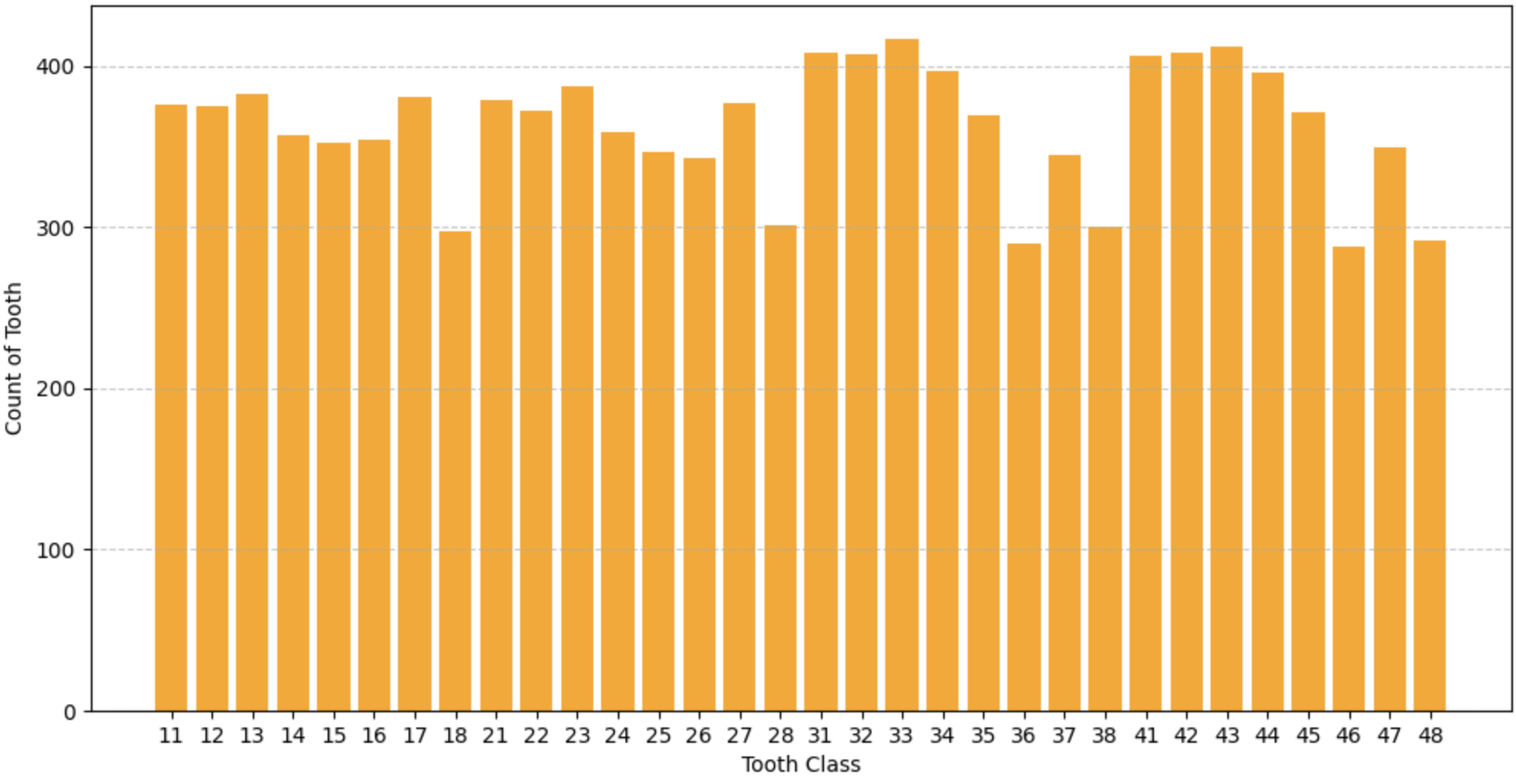}
        \caption{}
        \label{fig:subfigb}
    \end{subfigure}
    \hfill
    \caption{(a) Distribution of tooth counts across classes, showing balanced representation (b) Detailed count of each individual tooth (FDI notation 11–48), illustrating variability and coverage across the full dental arch}
    \label{fig:2}
\end{figure}

For the purpose of manual labeling, we utilized the semi-automated annotation tool Roboflow \cite{roboflow} to perform bounding box annotations required for the object detection task. Additionally, we employed the annotation tool Apeer \cite{apeer} to generate separate segmentation masks for each of the \(32\) teeth in the dataset. The annotation process was conducted by four students using a maximum-vote policy, whereby the annotation receiving the highest number of votes from the annotators was selected, followed by validation by an expert. These binary masks provided additional information by focusing on the fine contours and boundaries of the teeth. We converted the resulting segmented polygons into binary maps of size \((512 \times 512 \times 32)\)
. This comprehensive approach to annotation was crucial to ensure the success of our model in dental image analysis. The notation of the FDI World Dental Federation \cite{fdi2023}, illustrated in Figure ~\ref{fig:1}, is a standardized system to uniquely identify and label teeth. Widely adopted by dental professionals worldwide, it assigns a two-digit code to each tooth, ensuring clear and consistent communication of dental information across clinical and research contexts. Figure \ref{fig:2} shows that the dataset maintains balanced representation across all tooth classes and exhibits variability across different categories of panoramic X-ray images. This balance makes it well-suited for training tooth detection and segmentation models. Given its size and public availability, the dataset offers substantial potential for extensive use in tooth segmentation and detection tasks.
\begin{table}
  \centering
  \begin{tabular}{cccccc}
    \toprule
    Category & 32 Teeth & Restoration & Dental appliance & Images & Used Images \\
    \midrule
    1 & $\checkmark$ & $\checkmark$ &$\checkmark$ & 73 & 24\\
    2 & $\checkmark$ & $\checkmark$ &  & 220 & 72 \\
    3 & $\checkmark$ &  & $\checkmark$ & 45 & 15 \\
    4 & $\checkmark$ &  &  & 140 & 32 \\
    \midrule
    5 & &Images containing dental implant& & 120 & 37 \\
    6 & &Images containing more than 32 teeth& & 170 & 30 \\
    \midrule
    7 &  & $\checkmark$ & $\checkmark$ & 115 & 33 \\
    8 &  & $\checkmark$ &  & 457 & 140 \\
    9 &  &  & $\checkmark$ & 45 & 7 \\
    10 &  &  &  & 115 & 35 \\
    \midrule
    Total &  &  & & 1500 & 425\\
    \bottomrule
  \end{tabular}
  \vspace{5pt}
  \caption{Data Composition and Categorical Distribution of UFBA-UESC dataset.}
  \label{tab:2}
\end{table}
\subsection{Model Architecture}
\subsubsection{ Detection Module: YOLO-based Architecture}
\label{sec:yolo}
Currently, object detection is addressed using a range of models, predominantly categorized into single-stage and two-stage neural networks. Single-stage models, such as YOLO \cite{redmon2016you}, are known for their speed and efficiency compared to two-stage approaches. In this study, we employ YOLOv8\footnote{we have utilized the YOLOv8x version for all our training and testing purposes in this study.}, the most advanced version available at the time of experimentation. The main modules of YOLOv8 are listed below.
\begin{itemize}
\item \textbf{CSPDarknet53 Feature Extractor}: YOLOv8 uses CSPDarknet53, a variant of the Darknet architecture, as the feature extractor. This component comprises convolutional layers, batch normalization, and SiLU activation functions. The notable difference is that YOLOv8 replaces the original 6x6 convolutional layer with a \(3 \times 3\) convolutional layer to improve the extraction of characteristics.
\item \textbf{Module C2f}: YOLOv8 introduces the C2f module to enhance feature representation by efficiently combining high-level features with contextual information. Concatenates the output of bottleneck blocks, each comprising two \(3 \times 3\) convolutions with residual connections. Unlike YOLOv5’s C3 block, which includes an additional convolution layer, C2f reduces computational complexity. Used eight times throughout the architecture, this modification offers a notable efficiency gain.
\item \textbf{Detection head}: YOLOv8 adopts an anchor-free detection strategy that predicts object centers directly without predefined anchors. A key improvement is the use of different activation functions: a sigmoid function estimates objectness probability, while a softmax function predicts class probabilities. For optimization, YOLOv8 employs CIoU and DFL loss functions for bounding box regression and binary cross-entropy for classification, enhancing detection performance, particularly for small objects.
\end{itemize}
Although YOLOv8 is highly effective for object detection, it lacks pixel-level precision, making it less suitable for detailed segmentation tasks. It may struggle with closely packed or overlapping teeth and offers limited boundary accuracy due to its reliance on bounding boxes. These limitations motivate its integration with U-Net to achieve precise dental segmentation.
\subsubsection{Segmentation Module: U-Net-based Architecture}
Our methodology is based on the U-Net architecture \cite{ronneberger2015u}, a widely adopted model for semantic segmentation in medical imaging. U-Net features an encoder-decoder structure with symmetric skip connections: The encoder captures hierarchical features through convolutional layers, ReLU activations, and batch normalization, while max-pooling reduces spatial dimensions. The decoder reconstructs segmentation masks through transposed convolutions, progressively upsampling feature maps. Skip connections preserve spatial details and improve boundary accuracy. However, U-Net faces challenges in segmenting closely packed structures, such as individual teeth in panoramic X-rays, due to its reliance solely on feature extraction without incorporating explicit spatial priors. This can lead to over-segmentation or misclassification in complex regions. To overcome these issues, we propose OralBBNet, which integrates bounding box priors to improve localization and segmentation precision.
\subsubsection{OralBBNet Architecture}
OralBBNet builds on the U-Net framework by incorporating bounding box priors produced by the detection module into the learning process, enhancing segmentation precision through explicit spatial guidance. It's key innovation is the BB-Convolution layers, which refine feature maps using bounding-box-based supervision. While the encoder maintains U-Net’s hierarchical convolutional design, it integrates bounding-box-driven refinements at multiple levels. These bounding boxes act as spatial anchors, directing feature extraction to reduce errors in segmenting adjacent or overlapping structures.

\begin{figure}
     \centering
     \includegraphics[width=0.8\linewidth]{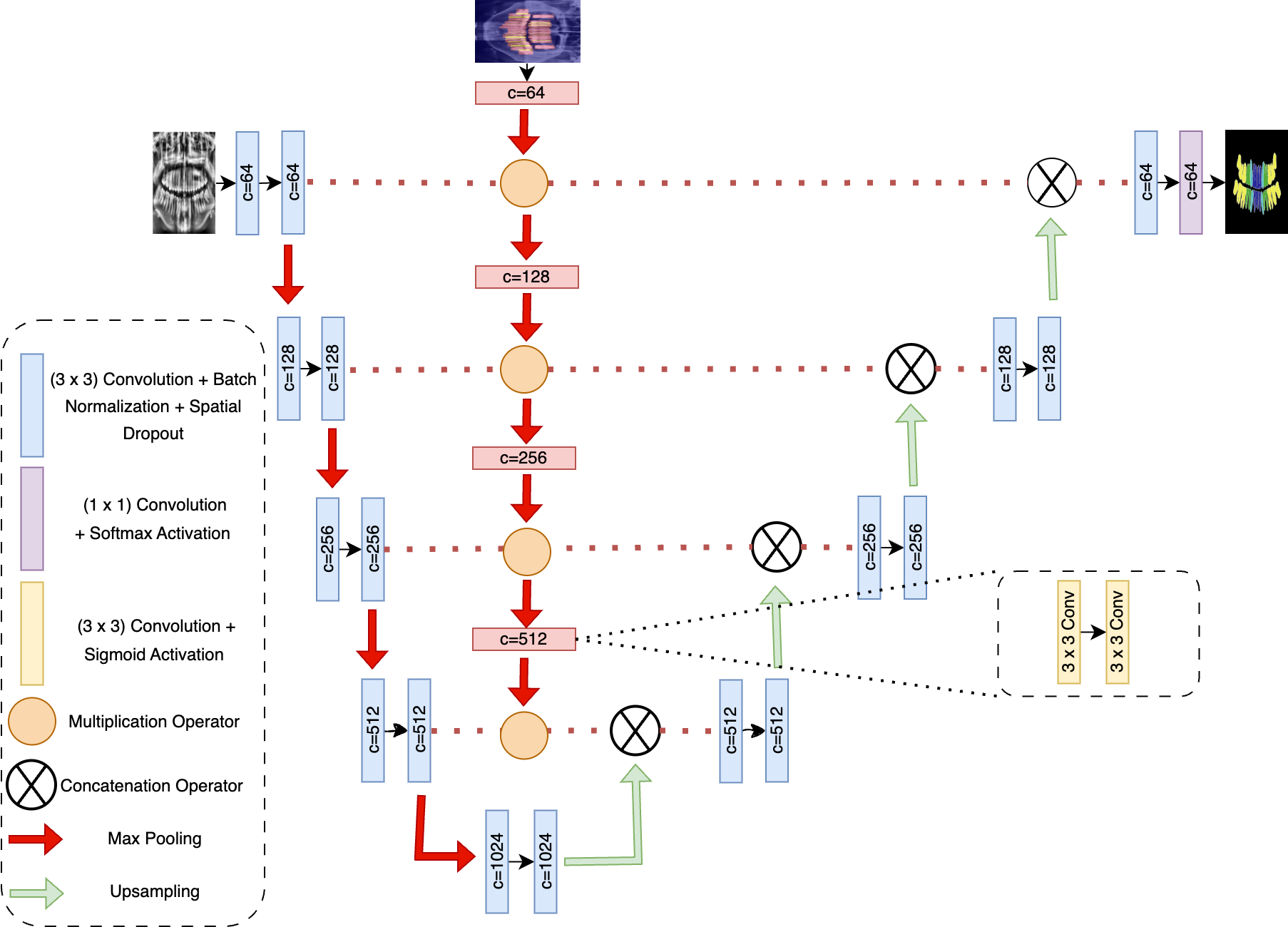}
     \caption{OralBBNet architecture integrating bounding box priors into U-Net for spatially guided dental image segmentation.}
     \label{fig:3}
 \end{figure}

Algorithm ~\ref{algo:1} outlines the processes for training, highlighting how spatial information from bounding boxes is utilized during both learning and prediction stages. As depicted in Figure ~\ref{fig:3}, BB-Convolution layers are inserted in the skip connections, featuring 2D max-pooling for reducing the bounding box map size, two convolution layers for spatial refinement, and a sigmoid activation to produce segmentation probability maps. These feature maps are combined element-wise with encoder outputs before being concatenated to decoder inputs, thereby incorporating localization information into the segmentation operation. The model concludes with a \((1 \times 1)\) convolution followed by a softmax activation, resulting in pixel-wise probability maps for accurate segmentation.

We use a regularized variant of Dice loss \cite{sudre2017generalised} to optimize the parameters of the OralBBNet during model training. Dice loss is a widely used metric in medical segmentation and computer vision tasks to calculate the similarity between two images.
\begin{equation}
    Loss = \frac{1}{N}\sum_{n=0}^{N}\frac{2\sum_{i=1}^{M}P_{n}(i)\hat{P_{n}}(i)}{\sum_{i=1}^{M}P_{n}(i)^{2} + \sum_{i=1}^{M}\hat{P_{n}}(i)^{2} } + \lambda\frac{1}{N}\sum_{n=0}^{N}\sum_{i=0}^{M}(P_{n}(i)-\hat{P_{n}}(i))^{2}
\end{equation}
Where $N$ is the number of class labels and $M$ is the number of pixels in each channel of the image. $P_{n}(i)$ and $\hat{P_{n}}(i)$ are the pixel values in the predicted map and ground truth label, respectively. Here $\lambda$ is a regularization constant. The latter component of the loss function plays a significant role in preventing overfitting, along with the spatial dropout layers introduced in the OralBBNet architecture.
\begin{algorithm}[H]
\caption{ OralBBNet Training Algorithm}
\footnotesize
\label{algo:1}
\scriptsize
\begin{algorithmic}
\State \textbf{Input:} X-ray Image $I \in \mathbb{R}^{H \times W \times C_i}$, Bounding Box Map $B \in \mathbb{R}^{H \times W \times C_b}$
\State \textbf{Output:} Segmented Image $S \in \mathbb{R}^{H \times W \times C_s}$
\State \textbf{Define ConvBlock:}
\State \hspace{1em} Input: Feature map $F$, filters $f$, kernel size $k$, drop rate $p$
\State \hspace{1em} $F \leftarrow \text{Conv}(F, f, k)$ \Comment{Convolution kernel}
\State \hspace{1em} $F \leftarrow (\text{ReLU})(F)$ \Comment{ReLU Activation}
\State \hspace{1em} $F \leftarrow \text{BatchNorm}(F)$ \Comment{Batch Normalization}
\State \hspace{1em} $F \leftarrow \text{SpatialDropout}(F, p)$ \Comment{Dropout}
\State \hspace{1em} Return $F$
\For{each epoch $t = 1, \dots, T$}
    \For{each sample $(I_i, B_i, G_i)$ in dataset $\mathcal{D}$}
        \State \textbf{Forward Pass:}
        \For{each level $x = 1, \dots, L$ in Encoder Layers}
            \State $F_x \leftarrow \text{ConvBlock}(F_{x-1}, f_x, 3, p)$
            \State $F_x \leftarrow \text{ConvBlock}(F_x, f_x, 3, p)$ 
            \State $F_x \leftarrow \text{MaxPool}(F_x, k=2)$ \Comment{Downsampling}
        \EndFor
        \If{$B_i \neq \emptyset$ (Bounding Box Prior Exists)}
            \For{each level $x = 1, \dots, L_{BB}$ in BB-Convolution Layers}
                \State $F_{BB} = \text{Maxpool}( B_i, k=2)$ 
                \State $F_{BB} = \text{Conv}( F_{BB}, f_x, 3)$
                \State $F_{BB} = \text{Conv}(F_{BB}, f_x, 3)$
                \State $F_{BB} = \sigma(W_{BB} * F_{BB} + b_{BB})$
                \State $F_x \leftarrow F_x \odot F_{BB}$ \Comment{Element-wise Multiplication with Encoder Output}
            \EndFor
        \EndIf
        \For{each level $x = L, \dots, 1$ in Decoder Layers}
            \State $F_{x} = \text{ConvTranspose}( F_{x+1}, s=2)$  \Comment{Upsampling}
            \State $F_{x}$ = \text{Concatenate BB-Convolution layer outputs with} $F_x$
            \State $F_x \leftarrow \text{ConvBlock}(F_x, f_x, 3, p)$
            \If{$x = 1$}
            \State $S_i \leftarrow \text{ConvBlock}(F_x, f_x, 1, p)$
            \Else
            \State $F_x \leftarrow \text{ConvBlock}(F_x, f_x, 3, p)$
            \EndIf
        \EndFor
        \State \textbf{Loss Computation:}
        \State $\mathcal{L}_{Dice} = 1 - \frac{2 \sum_{j} P_{ij} G_{ij}}{\sum_{j} P_{ij}^2 + \sum_{ij} G_{ij}^2} + \lambda \sum_{j} (P_{ij} - G_{ij})^2$ \Comment{Dice Loss with Regularization over pixels in} $S_i$
        \State \textbf{Backward Pass:}
        \State Compute Gradient $\frac{\partial \mathcal{L}}{\partial W} = \sum_j \left( \frac{\partial \mathcal{L}}{\partial P_{ij}} \cdot \frac{\partial P_{ij}}{\partial W} \right)$
        \State \textbf{Weight Update:}
        \State $W \leftarrow W - \eta \frac{\partial \mathcal{L}}{\partial W}$ \Comment{Gradient Descent (Adam)}
    \EndFor
    \State \Return Segmentation mask $S_i$
\EndFor
\end{algorithmic}
\end{algorithm}
\subsection{Training Pipeline and Hyperparameters Setup}
\begin{figure}
    \centering
    \includegraphics[width=1\linewidth]{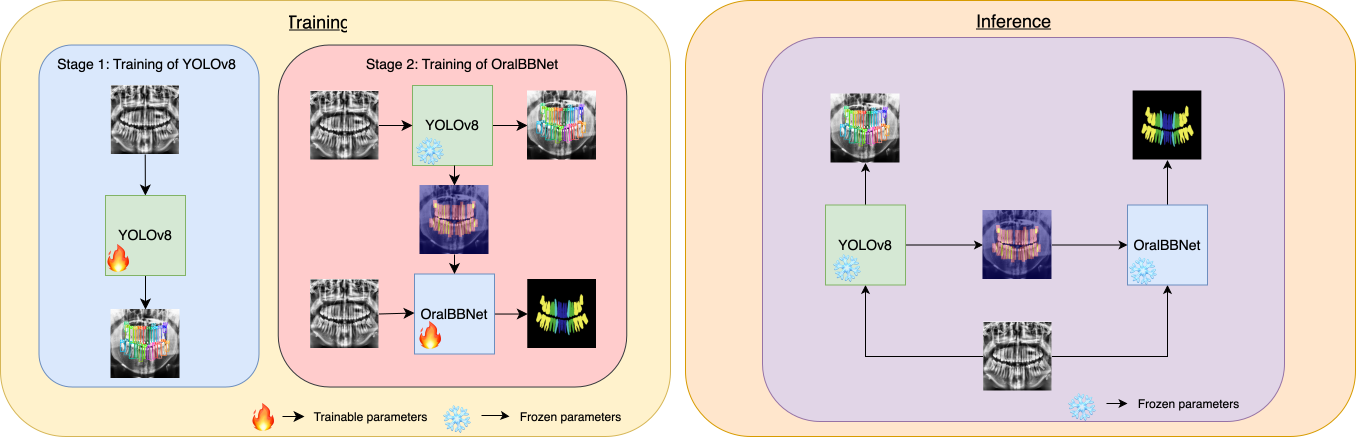}
    \caption{A two-stage pipeline where YOLOv8 is first trained for dental detection and then frozen to assist OralBBNet training for detailed tooth segmentation. In the inference phase, the panoramic x-ray is processed by YOLOv8 to obtain the teeth numbering and bounding boxes map, which, along with the x-ray, are then input into OralBBNet to produce the segmentation mask.}
    \label{fig:5}
\end{figure}

We propose a pipeline featuring a two-stage procedure that involves training the YOLOv8 and OralBBNet architectures independently, as illustrated in Figure ~\ref{fig:5}. In the first stage, the YOLOv8 detection module is trained to identify teeth, determine their positions, and extract bounding boxes. In the subsequent stage, the trained YOLOv8 model provides spatial prior information of bounding boxes, serving as a detection module while training OralBBNet for instance segmentation, ultimately producing segmentation masks as output. Furthermore, we ensured that the weights of YOLOv8 remained unchanged during the second stage of OralBBNet training to avoid collapse of weights in YOLOv8.
\subsubsection{Stage 1: Training of Detection Module}
For training YOLOv8, the images were resized to \(640 \times 640\), and histogram equalization was applied to enhance contrast. UFBA-425 dataset was expanded to \(1024\) panoramic x-rays using augmentation techniques such as random cropping \((0\%  \to 20\%\)) and brightness adjustment \((0\% \to 10\%\)). The dataset was split into \(894\) training images and \(128\) validation images and finally tested on UFBA-425.

\begin{figure}[ht]
    \centering
    \begin{subfigure}{0.43\linewidth}
        \centering
        \includegraphics[width=\linewidth]{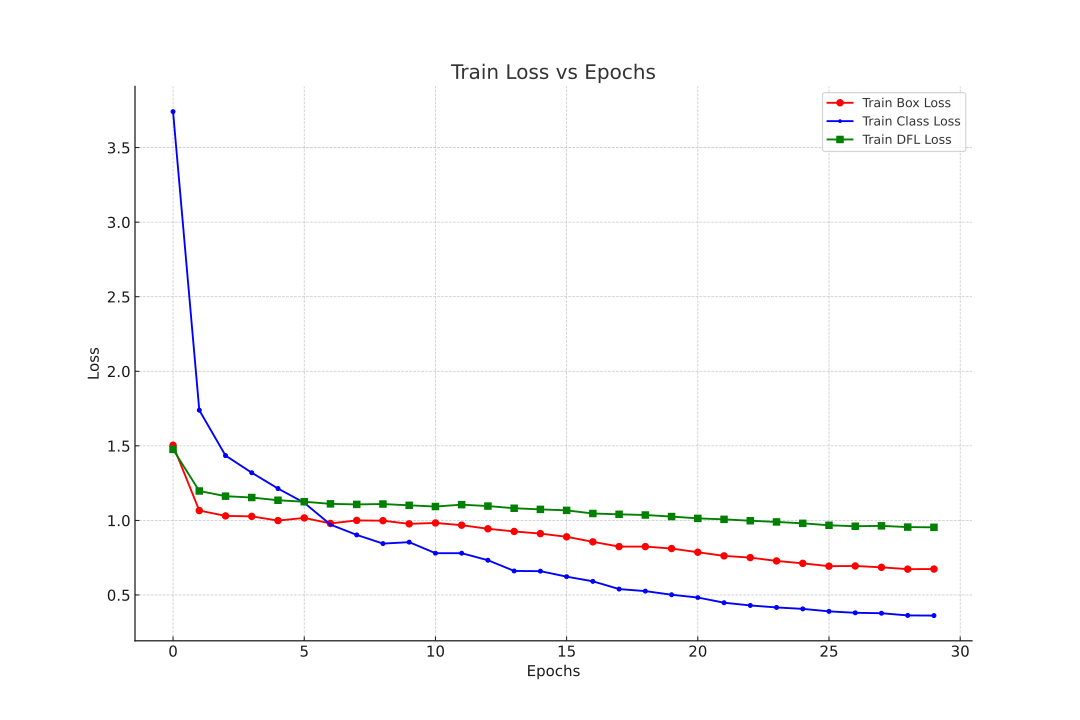}
        \caption{}
        \label{fig:subfiga1}
    \end{subfigure}
    \hfill
    \begin{subfigure}{0.5\linewidth}
        \centering
        \includegraphics[width=\linewidth]{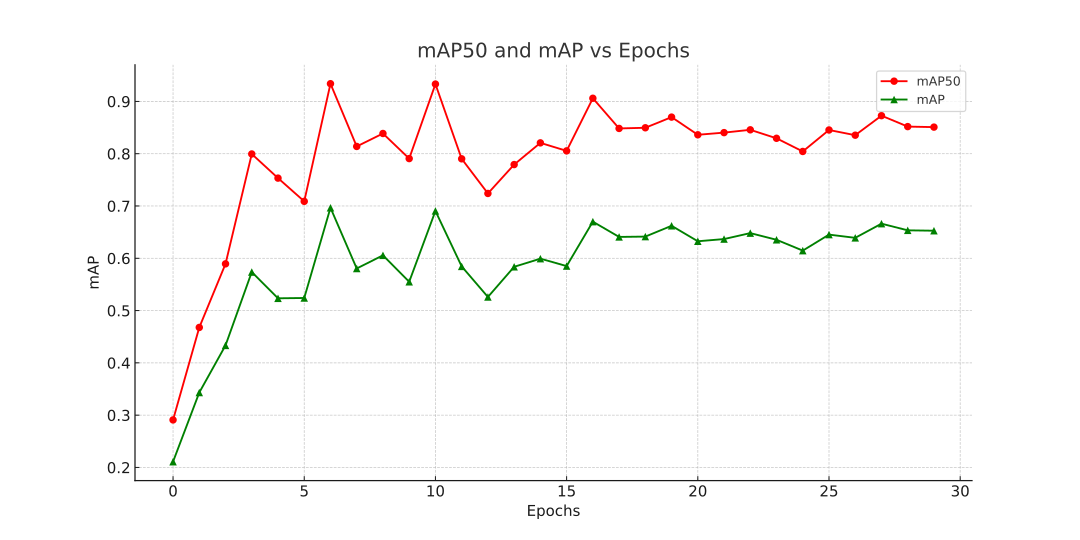}
        \caption{}
        \label{fig:subfigb1}
    \end{subfigure}
    \hfill
    \caption{(a) Training loss curves ( Box loss, DFL loss and Class loss) for YOLOv8 over training epochs (b) AP50 and mAP curves for YOLOv8 on validation dataset over training epochs}
    \label{fig:4}
\end{figure}

\textbf{Hyperparameter Setup: }The optimal results were achieved using an SGD optimizer with a learning rate of \(0.005\), a batch size of \(10\), a dropout rate of \(0.6\), and training over \(30\) epochs. Figure \ref{fig:4} illustrates the loss curves along with the mAP and AP50 scores on the validation dataset over the epochs. Notably, while the Box loss and DFL loss plateaued early, the classification loss continued to decrease throughout the training process. This trend suggests that the model underwent consistent and regularized training on the dataset.
\subsubsection{Stage 2: Training of Segmentation Module}
For training OralBBNet, the bounding box information from YOLOv8 was used to generate \(512\times512\times32\) bounding box binary maps, which served as spatial priors to BB-Convolution layers. \textit{Contrast Limited Adaptive Histogram Equalization} (CLAHE) \cite{clahe} with a contrast limit of \(0.02\) was applied to enhance panoramic x-ray image details. Horizontal and vertical flipping were used as data augmentation techniques, resulting in a training dataset of \(340\) images and a test dataset of \(85\) images.

\textbf{Hyperparameter Setup: }The optimal results were obtained with the Adam optimizer with a learning rate of \(0.0003\) with a momentum of 0.99, a batch size of \(2\) and dropout rate of \(0.12\) and regularization constant $\lambda$ of \(0.1\) and training over \(60\) epochs. The learning rate was halved if validation loss did not improve over a period of \(5\) epochs. The inference settings for YOLOv8 include a confidence threshold of \(0.5\) and an Intersection over Union (IoU) threshold of \(0.5\). An Nvidia A100 GPU with a 12-core CPU and 120GB RAM was used for all training and evaluation processes.
\begin{figure}[ht]
    \centering
    \begin{subfigure}{0.45\linewidth}
        \centering
        \includegraphics[width=\linewidth]{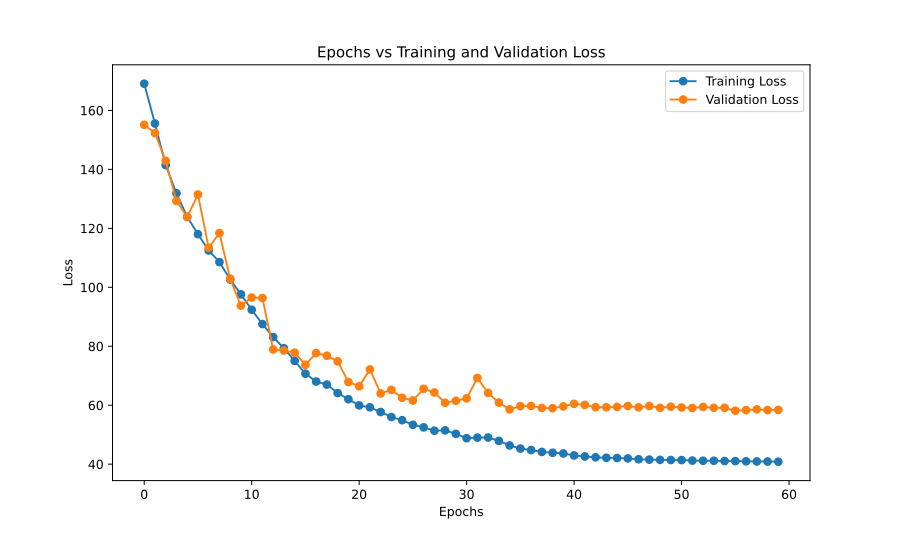}
        \caption{}
        \label{fig:subfiga2}
    \end{subfigure}
    \hfill
    \begin{subfigure}{0.45\linewidth}
        \centering
        \includegraphics[width=\linewidth]{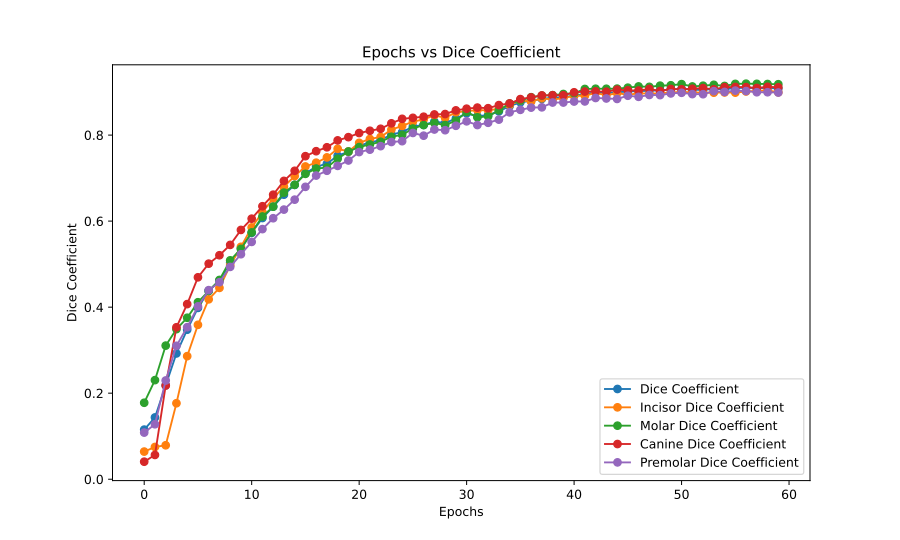}
        \caption{}
        \label{fig:subfigb2}
    \end{subfigure}
    \hfill
    \caption{(a) Training and validation loss curves for OralBBNet over training epochs (b) Categorical dice score curves on validation dataset of UFBA-425 for OralBBNet over training epochs}
    \label{fig:6}
\end{figure}
Figure ~\ref{fig:6} presents the training and validation loss curves over the epochs, along with the categorical dice score scores on the validation set.
\section{Experiments and Results}
\label{sec: results}
\subsection{Evaluation Metrics}
We have used several metrics to estimate the quality of the proposed model. We calculate the following metrics:
\begin{itemize}
    \item \emph{Accuracy} calculates the ratio of correctly predicted instances to the total number of instances.
    \begin{equation}\label{eq:acc}
    Accuracy = \frac{\sum_{i=1}^NM_{ii}}{\sum_{i=1}^N\sum_{j=1}^NM_{ij}}
    \end{equation}
    \item \emph{Recall}  measures the ability of the model to capture and correctly identify all relevant instances of a particular class.
    \begin{equation}\label{eq:recall}
    Recall_i = \frac{M_{ii}}{\sum_{j=1}^NM_{ij}}
    \end{equation}
    \item \emph{Precision}  measures the accuracy of the positive predictions made by the model. It indicates the proportion of true positives among all instances predicted as positive.
    \begin{equation}\label{eq:prec}
    Precision_i = \frac{M_{ii}}{\sum_{j=1}^NM_{ji}}
    \end{equation}
    \item \emph{mAP} calculates the precision-recall area's average under the curve for multiple classes at different confidence thresholds, providing a comprehensive evaluation of model performance.
    \begin{equation}\label{eq:mAP}
    mAP = \frac{1}{N}\sum_{i=1}^{N}AP_i = \frac{1}{N}\sum_{i=1}^{N}\sum_{j=1}^n(Recall_j - Recall_{j-1})Precision_j
    \end{equation}
    \item \emph{AP50} calculates the precision-recall area's average under the curve for multiple classes at a confidence threshold of \(0.5\).
\end{itemize}
In equations \ref{eq:acc}, \ref{eq:prec}, \ref{eq:recall} and \ref{eq:mAP}, $M_{ij}$ represents the corresponding element of a confusion matrix $M$ and $n$ is the number of thresholds and $N$ is the number of classes. mAP is used as the primary metric for teeth classification. In contrast, these metrics cannot provide a robust assessment of instance segmentation of teeth; the dice score ~\ref{eq:dice}  is used as the primary metric for instance segmentation of teeth. The dice score is a measure of the similarity between two sets, and it is used to quantify the agreement between the predicted segmentation masks and the ground truth masks. The dice score is defined as:
\begin{figure}
    \centering
    \includegraphics[width=0.9\linewidth]{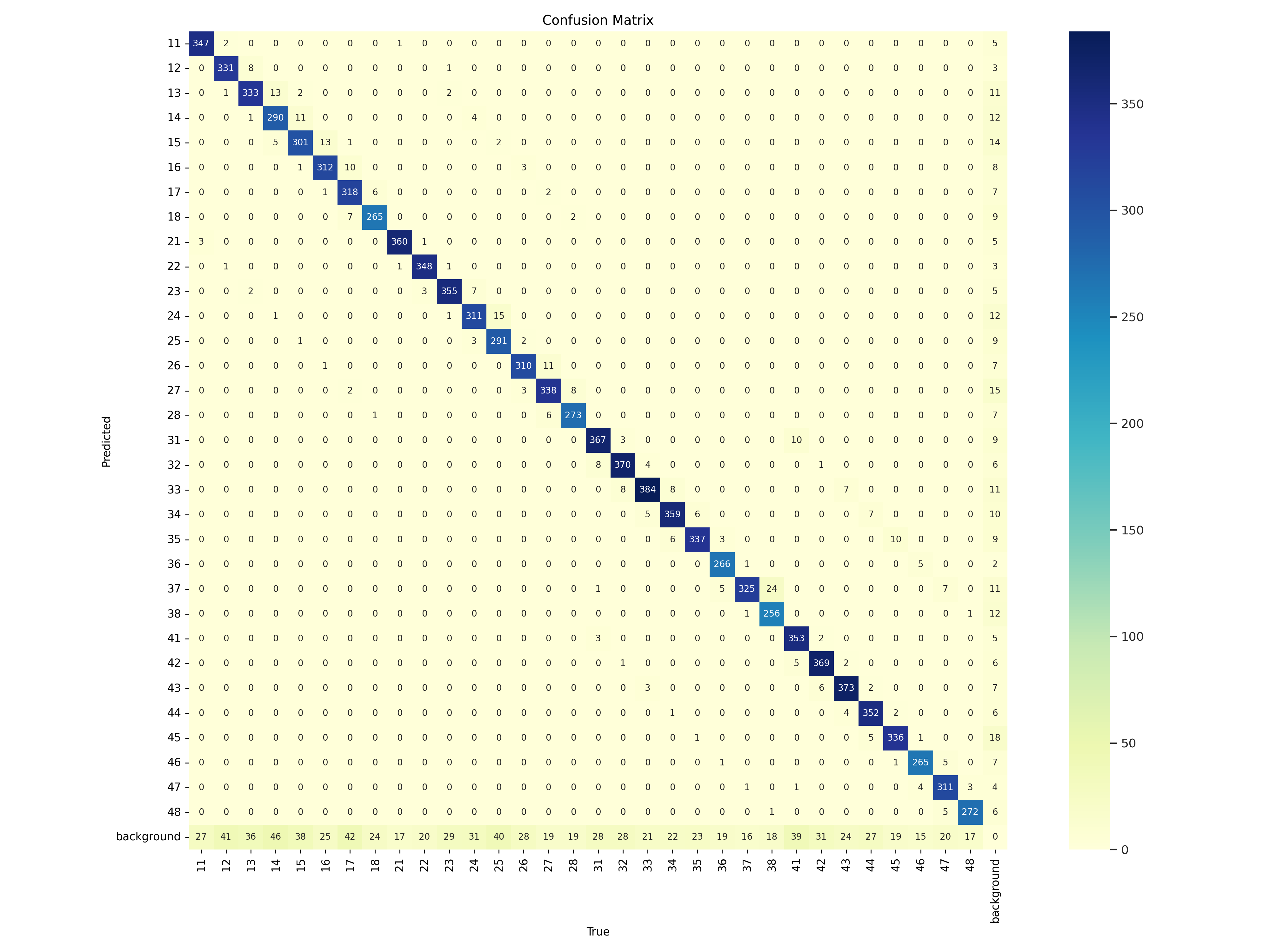}
    \caption{Confusion matrix for YOLOv8 predictions across all the 32 categories of teeth.}
    \label{fig:7}
\end{figure}
\begin{equation}\label{eq:dice}
    \text{Dice score} = \frac{1}{N} \sum_{n=0}^{N} \frac{2 \sum_{i=1}^{M} P_{n}(i) \hat{P_{n}}(i)}{\sum_{i=1}^{M} P_{n}(i)^{2} + \sum_{i=1}^{M} \hat{P_{n}}(i)^{2}}
\end{equation}
Where $N$ is the number of class labels and $M$ is the number of pixels in each channel of the image.\(P_{n}(i)\) and \(\hat{P_{n}}(i)\) are the pixel values in the predicted and the ground truth label, respectively.
\subsection{Evaluation of Detection Module: YOLOv8}
\subsubsection{Quantitative Analysis}
The performance of the detection head plays a critical role in the inference process of OralBBNet. Suboptimal performance in this component can significantly degrade segmentation outcomes, primarily due to the inadequate spatial priors being fed into OralBBNet. The YOLOv8 model was evaluated on the UFBA-425 dataset containing \(425\) images and \(11591\) instances of teeth and achieved a precision of \(94.30\), a recall of \(92.30\), and a mAP of \(74.90\) and a mean average precision at \(50\) (AP50) of \(94.60\) in all classes of teeth. 
The results indicated a striking balance between precision and recall and achieved an excellent mAP. These results suggest the effectiveness of the model in accurately localizing and recognizing teeth in the dataset. YOLOv8 performed better than Mask R-CNN, having a mAP of \(70.50\) \cite{Silva2020ASO} and other supervised and unsupervised methods \cite{Jader2018AutomaticST} and provided more accurate results. This demonstrates that a single-stage object detection algorithm can outperform a two-stage object detection method like Mask R-CNN. Due to the intricate intersection of the box region with other teeth in the image, the teeth belonging to the incisors have the lowest mAP score of all the tooth classes, with \(59.68\). YOLOv8 was able to localize the molar teeth precisely with a mAP of nearly \(78.55\) even though they have quite complex shapes, which shows the effectiveness of YOLOv8.
The confusion matrix in Figure \ref{fig:7} for YOLOv8 predictions for a confidence threshold of \(0.5\) and IOU threshold of \(0.5\), where the last row shows the number of instances of tooth being unclassified, indicates that YOLOv8 was able to accurately recognize the instances of every tooth with very few misclassifications but was unable to recognize some instances of every tooth above the confidence threshold; this indicates a lack of knowledge of instances of the tooth in some X-ray images of the dataset.
\subsubsection{Qualitative Analysis}
\begin{figure}
  \centering
  \begin{subfigure}{0.24\linewidth}
    \includegraphics[width=\linewidth]{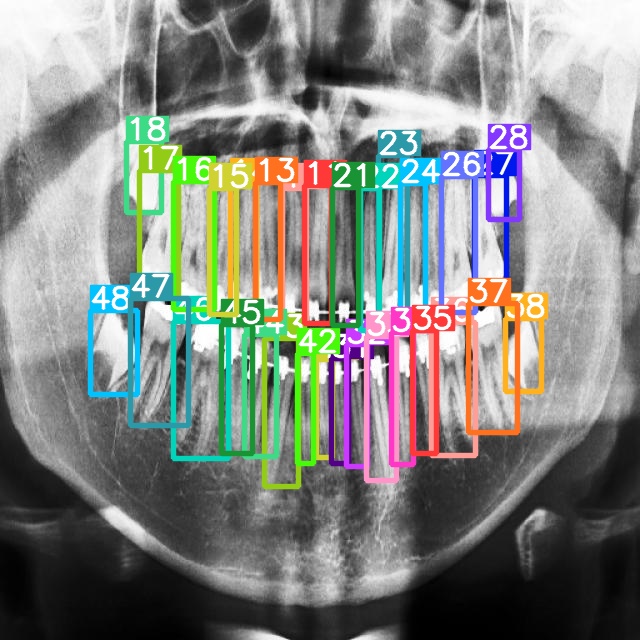}
    \label{fig:yolo_class_a}
  \end{subfigure}
  \begin{subfigure}{0.24\linewidth}
    \includegraphics[width=\linewidth]{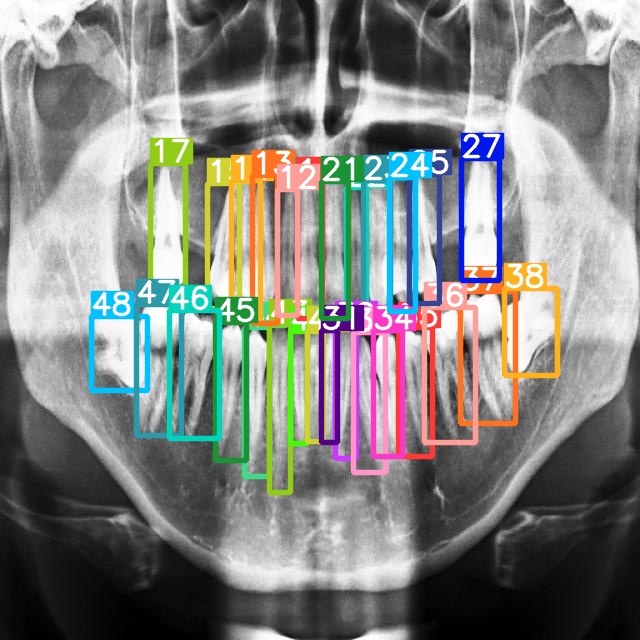}
    \label{fig:yolo_class_b}
  \end{subfigure}
  \begin{subfigure}{0.24\linewidth}
    \includegraphics[width=\linewidth]{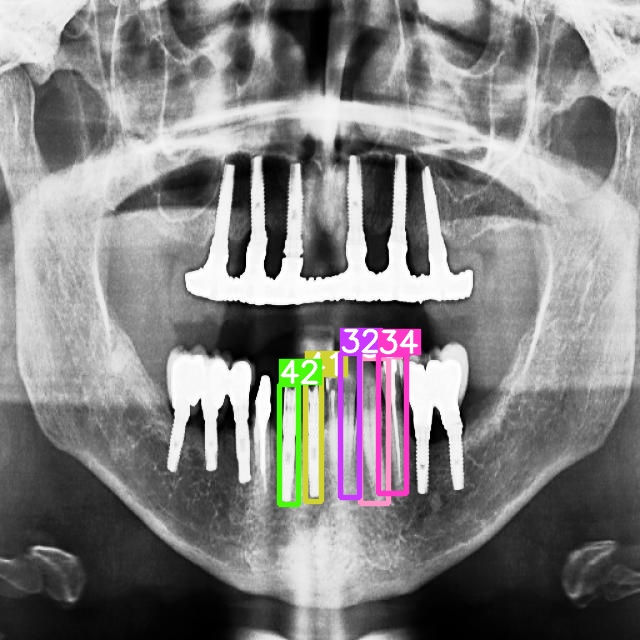}
    \label{fig:yolo_class_c}
  \end{subfigure}
  \begin{subfigure}{0.24\linewidth}
    \includegraphics[width=\linewidth]{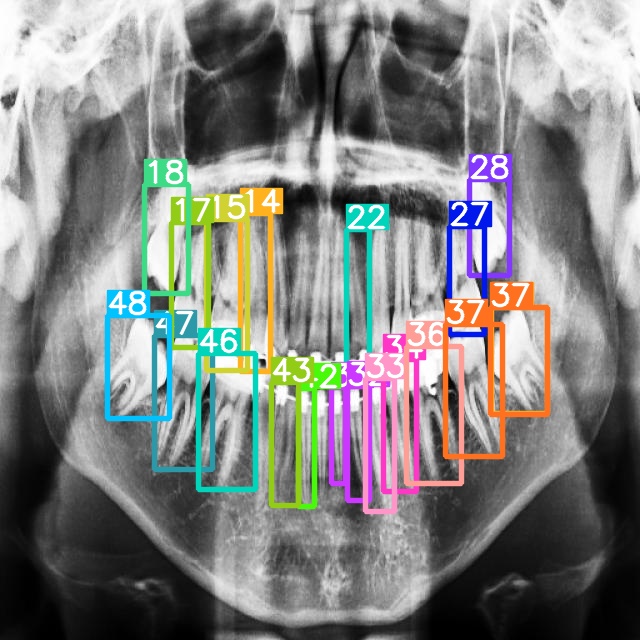}
    \label{fig:yolo_class_d}
  \end{subfigure}
  \begin{subfigure}{0.24\linewidth}
    \includegraphics[width=\linewidth]{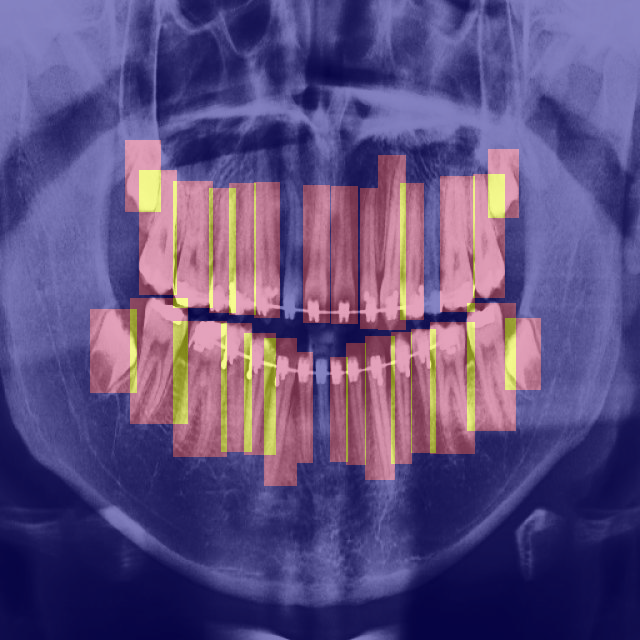}
    \caption{}
    \label{fig:yoloh_class_a}
  \end{subfigure}
  \begin{subfigure}{0.24\linewidth}
    \includegraphics[width=\linewidth]{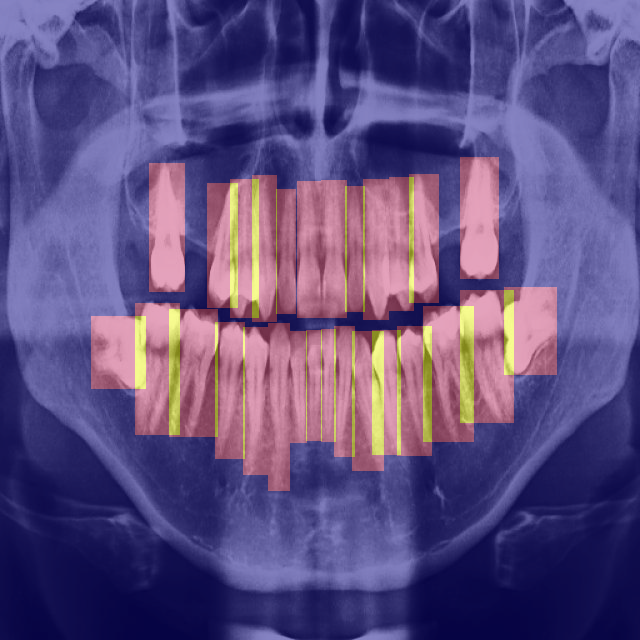}
    \caption{}
    \label{fig:yoloh_class_b}
  \end{subfigure}
  \begin{subfigure}{0.24\linewidth}
    \includegraphics[width=\linewidth]{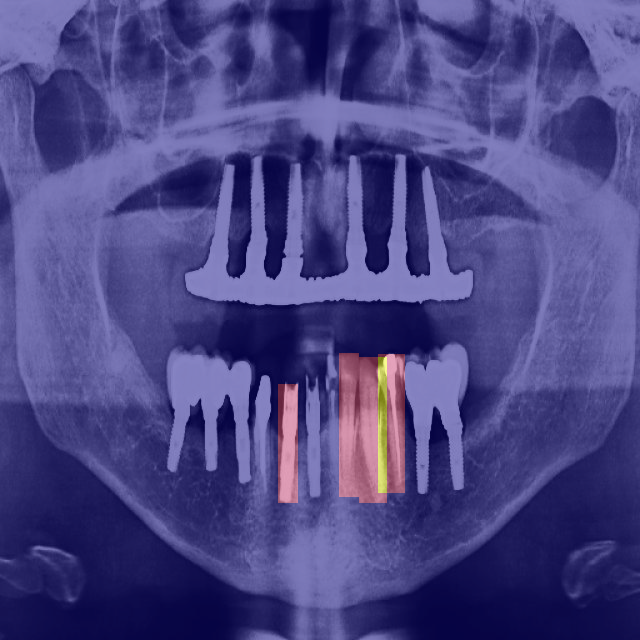}
    \caption{}
    \label{fig:yoloh_class_c}
  \end{subfigure}
  \begin{subfigure}{0.24\linewidth}
    \includegraphics[width=\linewidth]{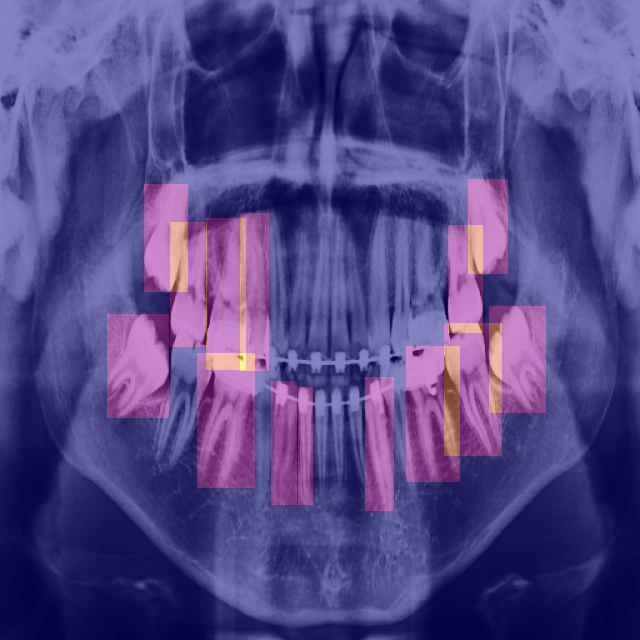}
    \caption{}
    \label{fig:yoloh_class_d}
  \end{subfigure}
\caption{Teeth numbering report with class labels and corresponding heatmaps of YOLOv8 results. Teeth numbering and bounding box heatmaps are of categories: (a)  category-1 image (b) category-8 image (c) category-5 image (d) category-6 image}
  \label{fig:yoloclass}
\end{figure}

The YOLOv8 model's performance analysis demonstrates accurate predictions for X-ray images lacking fillings and dental implants. However, challenges arise when encountering additional teeth, fillings, or implants. Notably, the scarcity of annotated data encompassing teeth with fillings or implants within the dataset contributes to this issue. Within Figure \ref{fig:yoloh_class_a}, Figure \ref{fig:yoloh_class_b}, the model presents precise predictions; conversely, Figure \ref{fig:yoloh_class_c}, Figure \ref{fig:yoloh_class_d} having missing tooth predictions and additional teeth, exhibits degraded performance. In cases involving fillings, implants, or decayed teeth, YOLOv8 demonstrates challenges in achieving precise detection.
\subsection{Evaluation of Segmentation Module: OralBBNet}
\subsubsection{Quantitative Analysis}
\begin{figure}
  \centering

  \begin{subfigure}{0.5\linewidth}
    \centering
    \includegraphics[width=\linewidth]{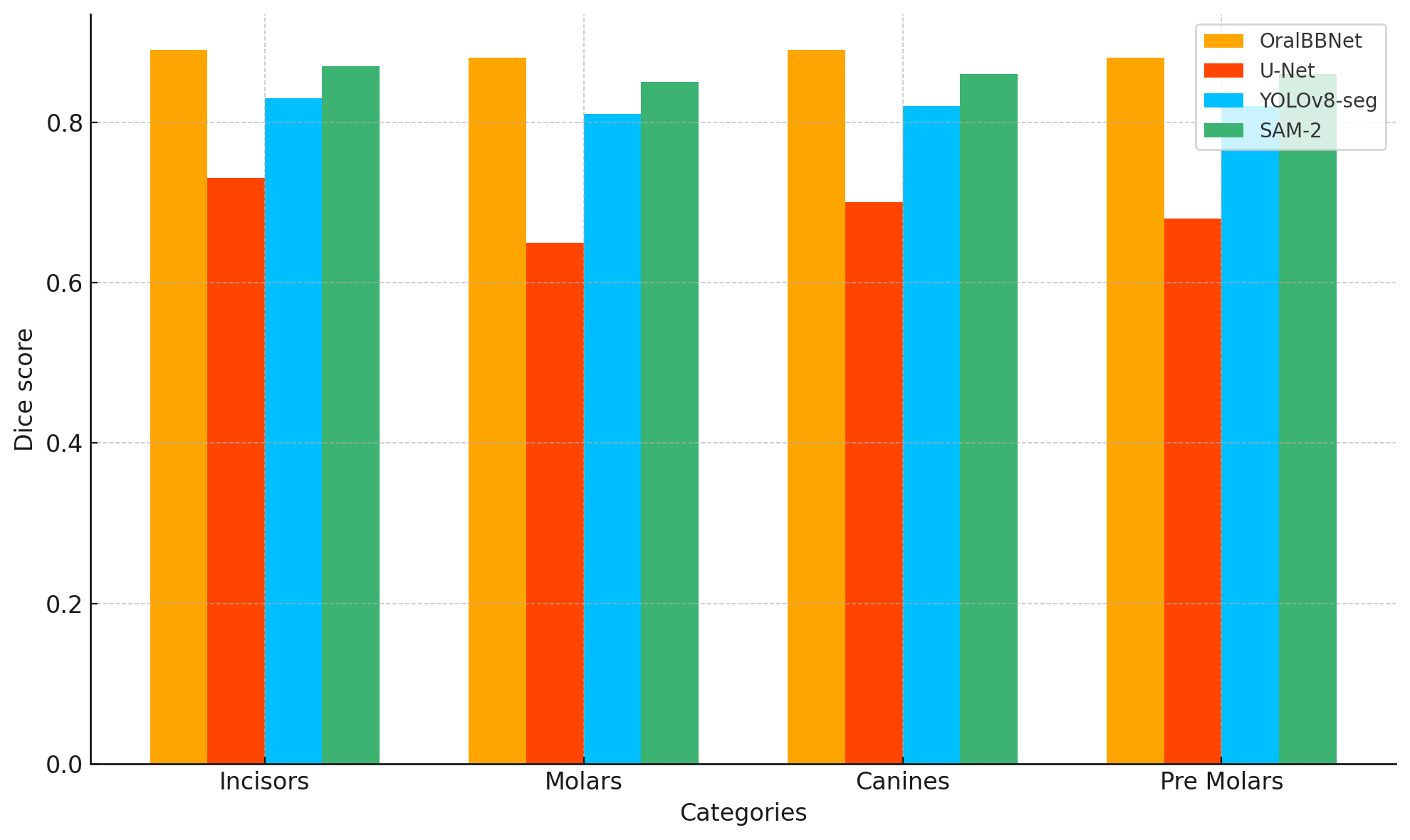}
    \caption{}
    \label{fig:comp1}
  \end{subfigure}%
  \hfill
  \begin{subfigure}{0.5\linewidth}
    \centering
    \includegraphics[width=\linewidth]{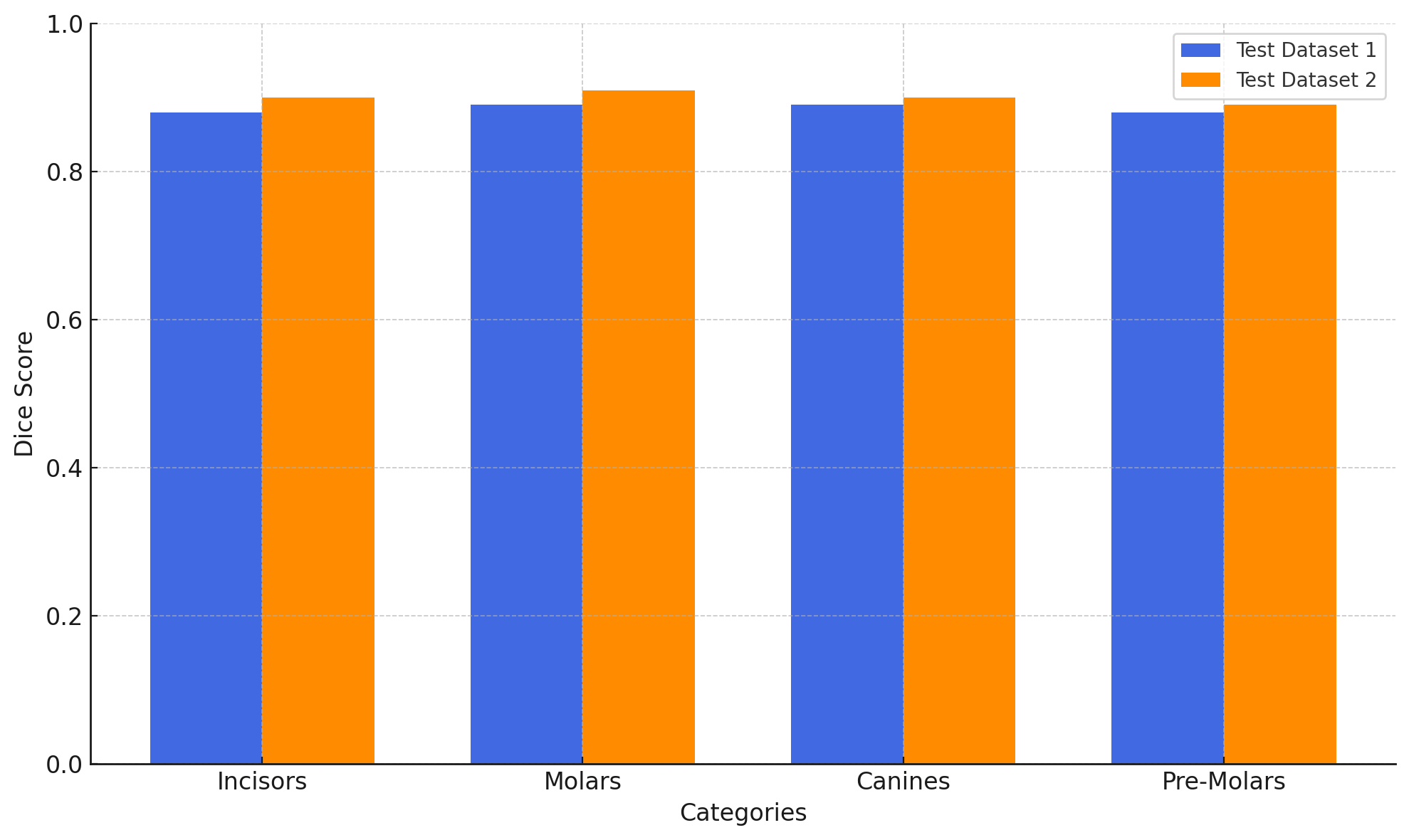}
    \caption{}
    \label{fig:comp2}
  \end{subfigure}

  \caption{(a) A comparison between performance of OralBBNet and other segmentation models trained on UFBA-425 train split (b) Segmentation performance of OralBBNet on test dataset 1 and test dataset 2 to understand the affects of critical X-ray images belonging to category-5 and category-6
}
  \label{fig:quantbbu}
\end{figure}

For evaluation of OralBBNet and U-Net, two test datasets were prepared from the original dataset. \textbf{Test Dataset 1}: contains 85 images belonging to all categories. \textbf{Test Dataset 2}: contains 72 images belonging to all categories except category-5 and category-6; i.e., X-ray images containing dental implants and X-ray images having more than 32 teeth are removed to differentiate and measure the effect on overall performance by more critical cases like category-5 and category-6 panoramic X-rays. Figure \ref{fig:comp1} shows that, when evaluated on test dataset 1 for all tooth kinds, OralBBNet performed better than U-Net with U-Net having an overall dice score of \(69.00\) and OralBBNet having an overall dice score of \(88.50\). U-Net showed the least dice scores on molars and pre-molars because of their complex geometries with more than two roots, U-Net had trouble precisely locating them. However, OralBBNet was able to do so because  spatial prior knowledge was incorporated from the detection head. For all tooth kinds, the dice score has increased by \(15\%\) to \(20\%\) when compared to U-Net, indicating the significance of  spatial prior knowledge in the instance segmentation of teeth. The primary limitation of OralBBNet was its incapacity to identify teeth in images with dental implants and X-ray images with more than \(32\) teeth because of the intricacy of the X-ray images and the presence of other dental instruments degraded the quality of prior knowledge and OralBBNet's performance. Figure \ref{fig:comp2} compares OralBBNet's performance on test dataset 1 and test dataset 2, having an overall dice score of \(88.50\) and \(89.80\), respectively. This indicates that OralBBNet's overall performance was affected by category-5 and category-6 X-ray images, resulting in scattered segmentation masks.
\subsubsection{Qualitative Analysis}
\begin{figure}
  \centering

  \begin{subfigure}{0.24\linewidth}
    \includegraphics[width=\linewidth]{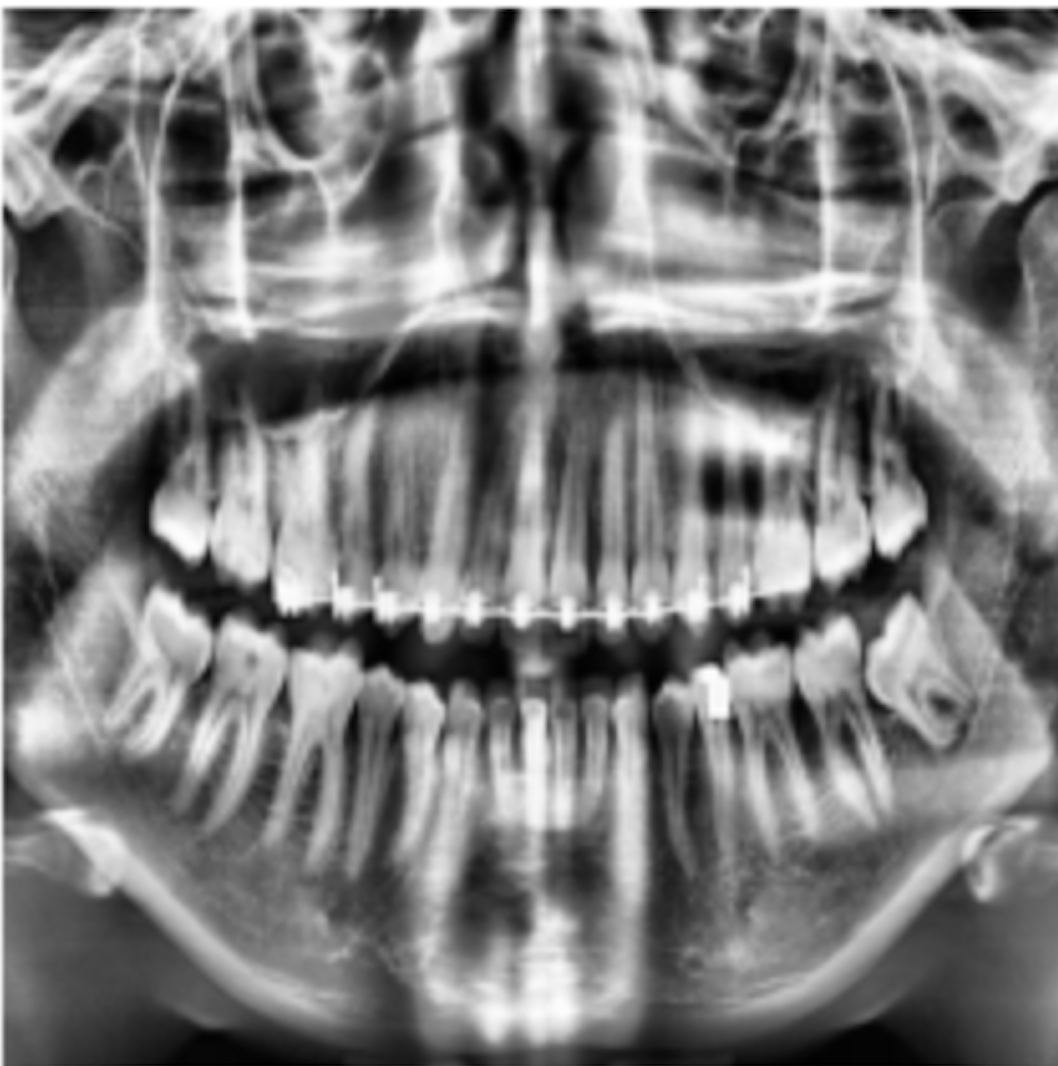}
  \end{subfigure}
  \begin{subfigure}{0.24\linewidth}
    \includegraphics[width=\linewidth]{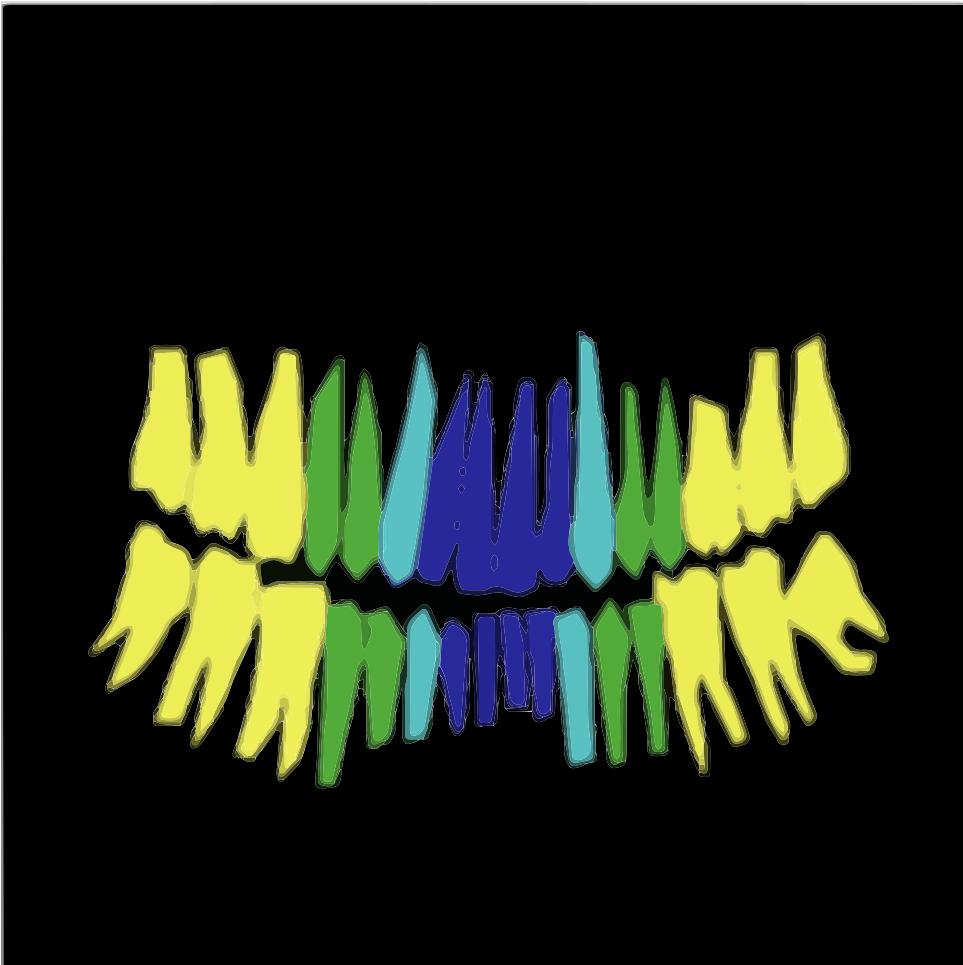}
  \end{subfigure}
  \begin{subfigure}{0.24\linewidth}
    \includegraphics[width=\linewidth]{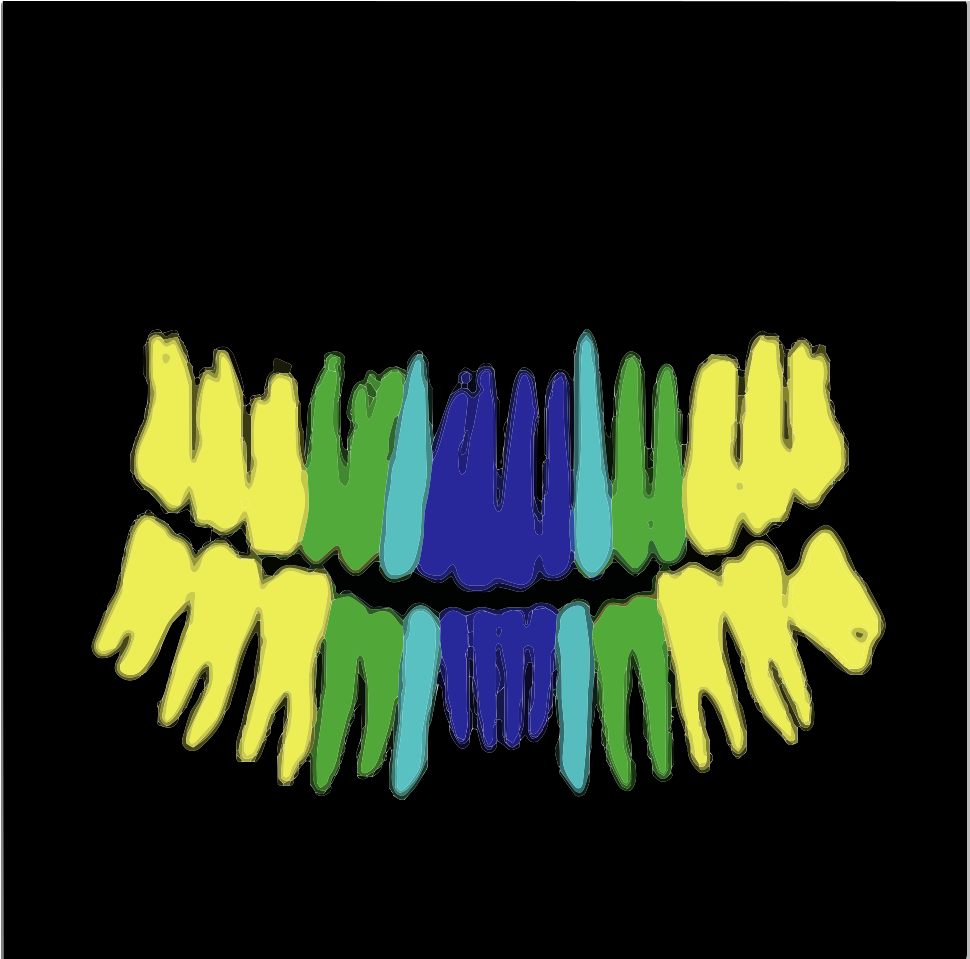}
  \end{subfigure}
  \begin{subfigure}{0.24\linewidth}
    \includegraphics[width=\linewidth]{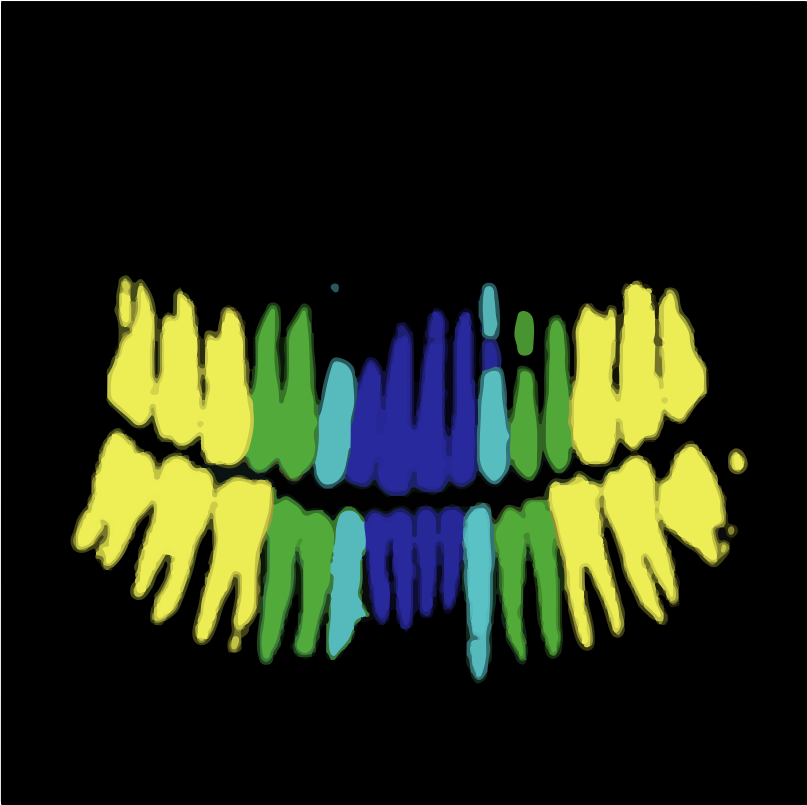}
  \end{subfigure}

  \medskip

  \begin{subfigure}{0.24\linewidth}
    \includegraphics[width=\linewidth]{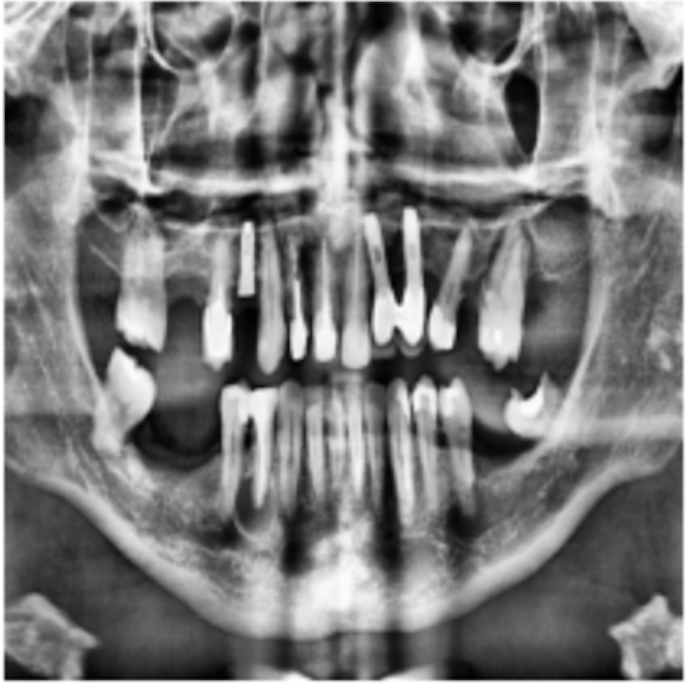}
    \label{fig:oralbbnet_a}
    \caption{Panoramic X-ray}
  \end{subfigure}
  \begin{subfigure}{0.24\linewidth}
    \includegraphics[width=\linewidth]{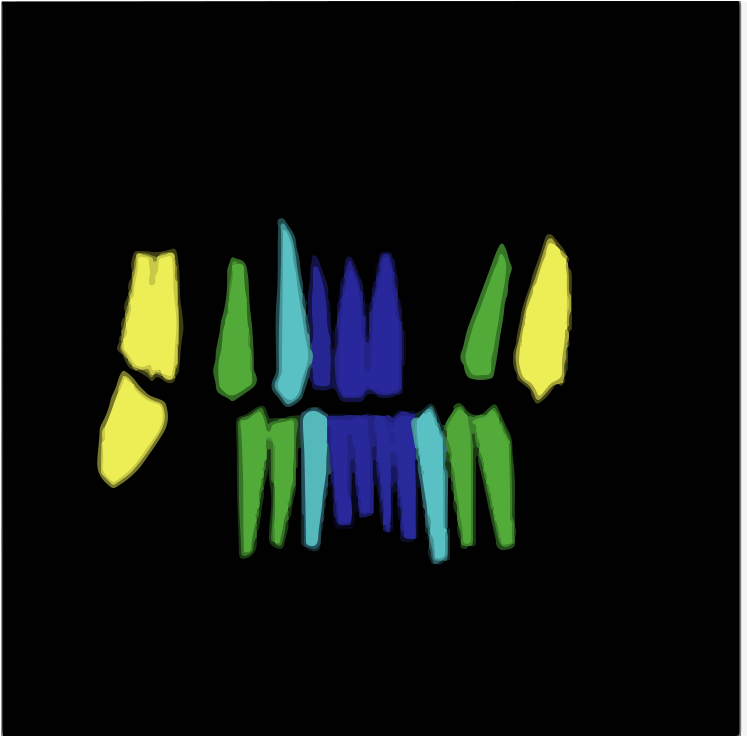}
    \label{fig:oralbbnet_b}
    \caption{Ground truth}
  \end{subfigure}
  \begin{subfigure}{0.24\linewidth}
    \includegraphics[width=\linewidth]{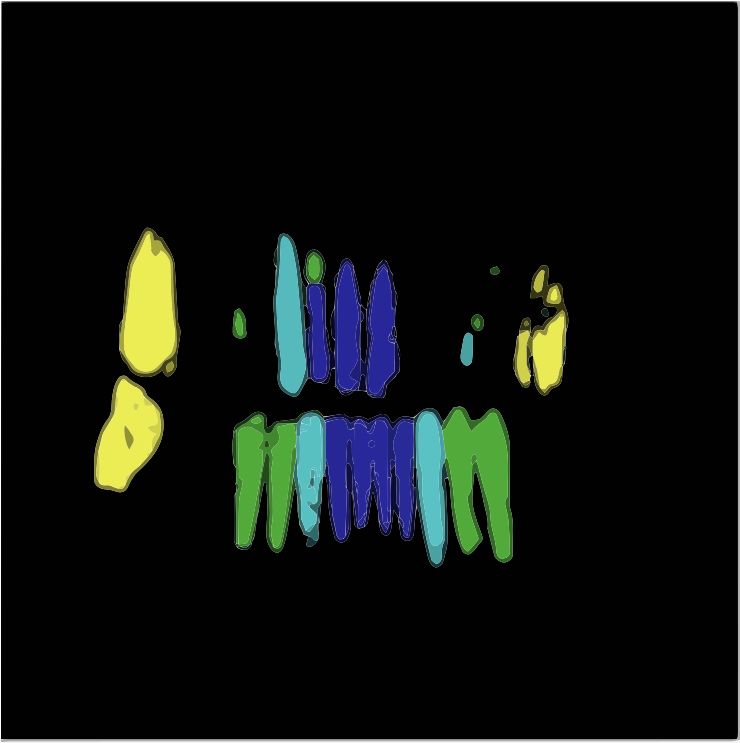}
    \label{fig:oralbbnet_c}
    \caption{OralBBNet}
  \end{subfigure}
  \begin{subfigure}{0.24\linewidth}
    \includegraphics[width=\linewidth]{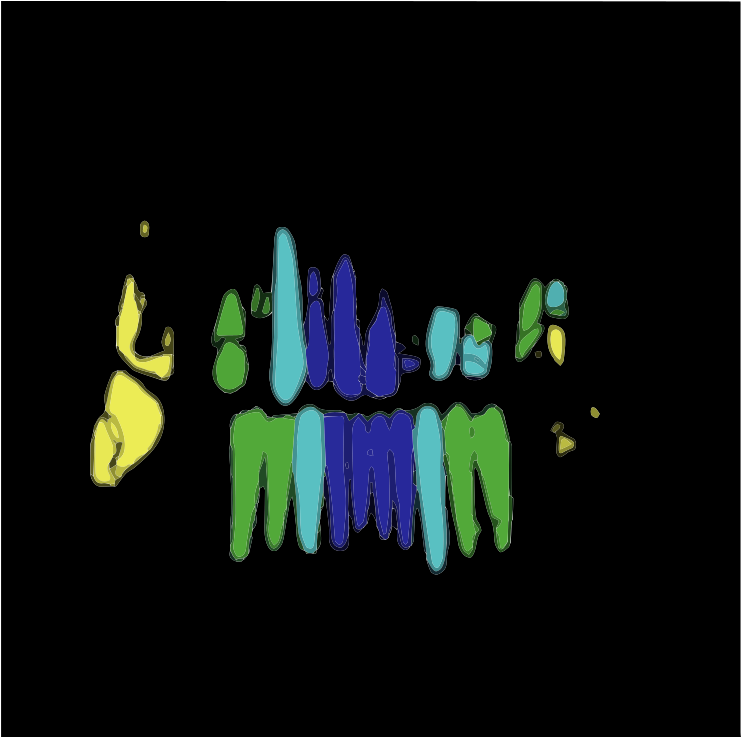}
    \label{fig:oralbbnet_d}
    \caption{U-Net}
  \end{subfigure}

  \caption{(top row) Superior Segmentation results of OralBBNet over U-Net for a category-1 panoramic X-ray (bottom row) Degraded results of both OralBBNet and U-Net for a category-5 panoramic X-ray with better results for OralBBNet compared to U-Net}
  \label{fig:bbu1}
\end{figure}

OralBBNet segmentation model notably outperforms the standard U-Net-based segmentation models from the point of quality. In Figure \ref{fig:bbu1}, the top row highlights the superior pixel-level classification of teeth achieved by OralBBNet, especially in complex tooth structures like molars and premolars, where U-Net encounters challenges. Nonetheless, the bottom row depicts the scenarios where bounding box predictions are degraded, OralBBNet's performance aligns with U-Net due to inadequate  spatial prior knowledge. Another observation was that OralBBNet successfully predicted most of the pixel densities even in category-5 and category-6 X-ray images but it struggled to accurately delineate the boundary pixels, which limits the dentists to extract precise spatial information from the segmentation masks.  Moreover, the dependence of OralBBNet's performance on bounding box predictions emphasizes the importance of comprehensive data annotation and robust performance of detection module.
\section{Comparison Studies}
\label{sec: ablations}
\subsection{Comparison of Detection Module with other Tooth Detection Models}
Table \ref{tab:yolodisc} compares the performance of the YOLOv8 architecture with other related dental detection models, evaluated on subsets of the UFBA-UESC dataset to ensure generality and reproducibility. This approach was necessary since the implementations of many existing models are not publicly available, making it difficult to benchmark their performance on the UFBA-425 subset. YOLOv8 outperformed Mask R-CNN, HTC, PANet and ResNetSt proposed by Silva et al. \cite{Silva2020ASO} on mAP score but has a lower AP50 score than these model architectures. Higher mAP score suggest that YOLOv8 is capable of detection with stricter IOU thresholds and the related models have utilized transfer learning for initializing the weights of the model architecture and have been evaluated on a smaller subset of UFBA-UESC dataset that does not contain complex panoramic X-rays belonging to category-5 and category-6, unlike YOLOv8 which was evaluated on a relatively more complex subset of UFBA-UESC dataset.
\subsection{Comparison of Segmentation Module with other Segmentation Models}
Table \ref{tab:bbunetdisc} summarizes the per-category and overall dice scores achieved by OralBBNet in comparison to U-Net, YOLOv8-seg, and SAM-2 \cite{ravi2024sam2} on the UFBA-425 dataset. OralBBNet attains the highest average dice score of \(88.50\) , surpassing SAM-2’s \(86.30\) and YOLOv8-seg’s \(82.00\). Across all four tooth types—incisors, canines, premolars, and molars—OralBBNet consistently delivers superior segmentation accuracy. This performance gain over other models can be attributed primarily to the spatial prior knowledge inherited from the YOLOv8 detection head, which guides OralBBNet’s network to more precise tooth boundary delineation.
\begin{table}[H]
\centering
\begin{minipage}{0.48\textwidth}
\centering
\begin{tabular}{c c c}
\hline
Model Architecture & mAP & AP50 \\
\hline
Mask R-CNN\cite{Silva2020ASO} & 70.5 & 97.2 \\
PANet\cite{Silva2020ASO} & 74.0 & 99.7 \\
HTC\cite{Silva2020ASO} & 71.1 & 97.3 \\
ResNeSt\cite{Silva2020ASO} & 72.1 & 96.8 \\
YOLOv8 \\(used in this study) & 74.9 & 94.6 \\
\hline
\end{tabular}
\vspace{5pt}
\caption{Comparison of teeth detection performance of                
 YOLOv8 and other related models evaluated on UFBA-UESC dataset.}
\label{tab:yolodisc}
\end{minipage}
\hfill
\begin{minipage}{0.48\textwidth}
\centering
\begin{tabular}{c c c c c}
\hline
\multicolumn{1}{c}{Model} & \multicolumn{4}{c}{Dice score} \\
\cline{2-5}
Architecture & Incisors & Canines & Premolars & Molars \\
\hline
U-Net & 73.29 & 69.92 & 67.62 & 64.98 \\
YOLOv8-seg & 82.78 & 81.91 & 81.89 & 81.42 \\
SAM-2      & 87.12 & 86.21 & 86.19 & 85.69 \\
OralBBNet\\ (proposed)  & 89.34 & 88.40 & 88.38 & 87.87 \\
\hline
\end{tabular}
\vspace{5pt}
\caption{Comparison of teeth segmentation performance of OralBBNet with other state-of-the-art models evaluated on UFBA-425.}
\label{tab:bbunetdisc}
\end{minipage}
\end{table}
\section{Limitations}
\label{sec: limitations}
\subsection{Limitations of UFBA-425 Dataset}
The UFBA-425 dataset offers a valuable foundation for dental segmentation and numbering, but it has notable limitations in scope and depth of annotation. Although the complete collection comprises \(1500\) panoramic X-rays that span ten clinical categories, only \(425\) images were manually annotated for both tooth detection and instance segmentation, limiting the effective training set to less than a third of the total data. Within this annotated subset, complex cases remain under-represented: only \(37\) images contain dental implants and only \(30\) feature supernumerary teeth, which means the model sees very few examples of these critical clinical scenarios during training. This sparsity risks overfitting to the more common cases and hampers generalization to the full diversity of real-world dental X-rays.
\subsection{Limitations of OralBBNet}
OralBBNet’s architecture significantly improves segmentation by injecting YOLOv8’s bounding-box priors into a U-Net backbone, but this tight coupling introduces its own vulnerabilities. Whenever YOLOv8 fails to localize a tooth particularly in implants or (> 32-tooth) images, the BB-Convolution layers receive noisy spatial guidance, and the network performance degrades, producing scattered or missing masks. Furthermore, the pipeline trains YOLOv8 first and then \textit{freezes} it before training OralBBNet, preventing any feedback from the segmentation stage from refining the detector but crucial to ensure collapse of weights. This lack of end-to-end co-adaptation means that detection errors propagate uncorrected into the segmentation.

\section{Conclusion}
\label{sec: conclusion}
Our research demonstrated promising progress in segmentation performance by integrating prior spatial information into the OralBBNet skip connections. The UFBA-425 dataset aids in advancing research in deep learning techniques for dental segmentation and detection, yet there is still considerable room for improvement since even OralBBNet and cutting-edge models like SAM-2 have not yet achieved optimal performance. However, despite these strides, several areas for improvement were identified. The detection head faced challenges in accurately predicting numerous labels belonging to images of categories 5 and 6 within the dataset that adversely affected overall model performance. Addressing these imbalances by expanding the size of the dataset rather than enhancing strategies could boost performance. We conclude that expanding the dataset and improving the detection head could potentially lead to enhanced segmentation and classification results in upcoming research projects.

\section{Data Availability}
\label{sec: dataset link}
The dataset and the source code for the pipelines are available at \href{https://github.com/devichand579/Instance_seg_teeth}{OralBBNet} or on request from the authors.
\section{Declaration of interests}
The authors declare that they have no known competing financial interests or personal relationships that could have appeared to influence the work reported in this paper.

\bibliographystyle{unsrt}
\bibliography{references} 
\end{document}